\documentclass[a4paper,UKenglish]{article}

\usepackage[utf8]{inputenc}
\usepackage{microtype}
\usepackage[left=3.4cm,right=3.4cm,bottom=3.5cm,top=3.4cm] {geometry}
\usepackage{graphicx}
\usepackage{hyperref}
\usepackage[numbers]{natbib}
\newcommand{\footremember}[2]{%
    \footnote{#2}
    \newcounter{#1}
    \setcounter{#1}{\value{footnote}}%
}
\newcommand{\footrecall}[1]{%
    \footnotemark[\value{#1}]%
} 

\newcommand{\abstracttitle}[1]{\subsubsection{#1}}
\newcommand{\abstractauthor}[2][None]{\noindent\emph{#1}\vspace{0.25cm}}
\newcommand{\license}{}
\newcommand{\jointwork}[1]{Joint work with \emph{#1}.}
\newcommand{\abstractref}[1]{}
\newcommand{\abstractrefurl}[1]{}

\renewcommand{\cite}[1]{\citep{#1}}
\title{AI for Science: An Emerging Agenda}

\author{%
  Philipp Berens\footremember{tuebingen}{Universität Tübingen, DE}%
  \and Kyle Cranmer\footremember{wisconsin}{University of Wisconsin - Madison, USA}%
  \and Neil D. Lawrence\footremember{cambridge}{University of Cambridge, UK}
  \and Ulrike von Luxburg\footrecall{tuebingen}
  \and Jessica Montgomery\footrecall{cambridge}
  }

\begin{document}

\maketitle

\begin{abstract}
This report documents the programme and the outcomes of Dagstuhl Seminar 22382 "Machine Learning for Science: Bridging Data-Driven and Mechanistic Modelling".

Today's scientific challenges are characterised by complexity.
Interconnected natural, technological, and human systems are influenced
by forces acting across time- and spatial-scales, resulting in complex
interactions and emergent behaviours. Understanding these phenomena ---
and leveraging scientific advances to deliver innovative solutions to
improve society's health, wealth, and well-being --- requires new ways of
analysing complex systems.

The transformative potential of AI stems from its widespread
applicability across disciplines, and will only be achieved through
integration across research domains. AI for science is a rendezvous
point. It brings together expertise from AI and application domains;
combines modelling knowledge with engineering know-how; and relies on
collaboration across disciplines and between humans and machines.
Alongside technical advances, the next wave of progress in the field
will come from building a community of machine learning researchers,
domain experts, citizen scientists, and engineers working together to
design and deploy effective AI tools.

This report summarises the discussions from the seminar and provides a roadmap to suggest how different communities can collaborate to deliver a new wave of progress in AI and its application for scientific discovery.
\end{abstract}

\section*{Summary}

Today's scientific challenges are characterised by complexity.
Interconnected natural, technological, and human systems are influenced
by forces acting across time- and spatial-scales, resulting in complex
interactions and emergent behaviours. Understanding these phenomena ---
and leveraging scientific advances to deliver innovative solutions to
improve society's health, wealth, and well-being --- requires new ways of
analysing complex systems.

Artificial intelligence (AI) offers a set of tools to help make sense of
this complexity. In an environment where more data is available from
more sources than ever before --- and at scales from the atomic to the
astronomical --- the analytical tools provided by recent advances in AI
could play an important role in unlocking a new wave of research and
innovation. The term AI today describes a collection of tools and
methods, which replicate aspects of intelligence in computer systems.
Many recent advances in the field stem from progress in machine
learning, an approach to AI in which computer systems learn how to
perform a task, based on data.

Signals of the potential for AI in science can already be seen in many
domains. AI has been deployed in climate science to investigate how
Earth's systems are responding to climate change; in agricultural
science to monitor animal health; in development studies, to support
communities to manage local resources more effectively; in astrophysics
to understand the properties of black holes, dark matter, and
exoplanets; and in developmental biology to map pathways of cellular
development from genes to organs. These successes illustrate the wider
advances that AI could enable in science. In so doing, these
applications also offer insights into the science of AI, suggesting
pathways to understand the nature of intelligence and the learning
strategies that can deliver intelligent behaviour in computer systems.

Further progress will require a new generation of AI models. AI for
science calls for modelling approaches that can: facilitate
sophisticated simulations of natural, physical, or social systems,
enabling researchers to use data to interrogate the forces that shape
such systems; untangle complicated cause-effect relationships by
combining the ability to learn from data with structured knowledge of
the world; and work adaptively with domain experts, assisting them in
the lab and connecting data-derived insights to pre-existing domain
knowledge. Creating these models will disrupt traditional divides
between disciplines and between data-driven and mechanistic modelling.

The roadmap presented here suggests how these different communities can
collaborate to deliver a new wave of progress in AI and its application
for scientific discovery. By coalescing around the shared challenges for
AI in science, the research community can accelerate technical progress,
while deploying tools that tackle real-world challenges. By creating
user-friendly toolkits, and implementing best practices in software and
data engineering, researchers can support wider adoption of effective AI
methods. By investing in people working at the interface of AI and
science -- through skills-building, convening, and support for
interdisciplinary collaborations --- research institutions can encourage
talented researchers to develop and adopt new AI for science methods. By
contributing to a community of research and practice, individual
researchers and institutions can help share insights and expand the pool
of researchers working at the interface of AI and science. Together,
these actions can drive a paradigm shift in science, enabling progress
in AI and unlocking a new wave of AI-enabled innovations.

The transformative potential of AI stems from its widespread
applicability across disciplines, and will only be achieved through
integration across research domains. AI for science is a rendezvous
point. It brings together expertise from AI and application domains;
combines modelling knowledge with engineering know-how; and relies on
collaboration across disciplines and between humans and machines.
Alongside technical advances, the next wave of progress in the field
will come from building a community of machine learning researchers,
domain experts, citizen scientists, and engineers working together to
design and deploy effective AI tools.

\section{Introduction: bridging data driven and mechanistic
modelling}\label{introduction-bridging-data-driven-and-mechanistic-modelling}

The 21\textsuperscript{st} century has been characterised as the century
of complexity.\footnote{This quote is attributed to Stephen Hawking, in
  an interview with the San Jose Mercury News in January 2000.} Shifting
social, economic, environmental, and technological forces have created
increasingly interconnected communities, affected by `wicked' problems
in domains such as health, climate, and economics \cite{Rittel-dilemmas73}. This complexity is reflected in today's scientific
agenda: whether in natural, physical, medical, environmental, or social
sciences, researchers are often interested in the dynamics of complex
systems and the phenomena that emerge from them.

Science has always proceeded through the collection of data. Through
their experiments and observations, researchers collect data about the
world, use this data to develop models or theories of how the world
works, make predictions from those models, then test those predictions,
leading to further refinements to the model and the underpinning theory.
Digitisation of daily activities---in the lab, and elsewhere -- means
that researchers today have access to more data from a greater range of
sources than ever before. In parallel, more sophisticated tools to
collect data have opened new scales of scientific inquiry, from detailed
patterns of gene expression to light signals from other galaxies. Data
proliferation is both a signal of the complexity of today's environment,
and an opportunity to make sense of such complexity.

Advances in artificial intelligence (AI) have produced new analytical
tools to make sense of these data sources. The term `AI' today describes
a collection of methods and approaches to create computer systems that
can perform tasks that would typically be associated with `intelligent'
behaviour in living systems.\footnote{While not the only branch of the
  field, machine learning is the approach to AI that has delivered many
  of the recent advances in AI. Machine learning is an approach to AI in
  which models process data, learning from that data to identify
  patterns or make predictions. In this document, the terms machine
  learning and AI are used interchangeably.} In this document, the term
AI is used broadly, to refer to algorithmic decision-making systems that
combine data, mathematical models, and compute power to make predictions
about the world.

AI is already unlocking progress across research disciplines:

\begin{itemize}
\item
  In Earth sciences, it is helping researchers investigate how different
  parts of the Earth's biosphere interact, and are affected by climate
  change.\footnote{These examples are inspired by talks given at the
    Dagstuhl seminar; these are provided later in the document. This
    example is inspired by Markus Reichstein's talk.}
\item
  In climate science, it supports modelling efforts to reconstruct
  historical climate patterns, enabling more accurate predictions of
  future climate variability.\footnote{This example is inspired by Ieva
    Kazlauskaitė's talk.}
\item
  In agricultural science, it is helping farmers access faster diagnoses
  of animal diseases, enabling more effective responses.\footnote{This
    example is inspired by Dina Machuve's talk.}
\item
  In astrophysics, it is advancing understandings of the nature of dark
  matter and its role in the Universe.\footnote{This example is inspired
    by Siddharth Mishra-Sharma's talk.}
\item
  In developmental biology, it is generating insights into the genetic
  processes that shape how cells develop and differentiate into
  specialist roles.\footnote{This example is inspired by Maren Büttner's
    talk.}
\item
  In environmental science, it allows researchers to analyse the
  features of natural environments more accurately, aiding land and
  resource managers.\footnote{This example is inspired by Christian
    Igel's talk.}
\item
  In neuroscience, it can help model how different neural circuits fire
  to deliver different behaviours in animals.\footnote{This example is
    inspired by Jakob Macke's talk.}
\end{itemize}

The diversity of these successes illustrates the transformative
potential of AI for research across the natural, physical, social,
medical, and computer sciences, arts, humanities, and engineering. By
enabling researchers to extract insights from a greater volume of data,
drawn from a wider variety of sources, and operating across multiple
dimensions and scales, AI could unlock new understandings of the world.
In so doing, AI could influence the conduct of science itself.
AI-enabled analytical tools mean researchers can now generate
sophisticated simulations of natural or physical systems, creating
`digital siblings' of real-world systems that can be used for
experimentation and analysis. Machine learning models that combine the
ability to learn adaptively from data with the ability to make
structured predictions reflecting the laws of nature can help
researchers untangle the web of cause-effect relationships that drive
the dynamics of complex systems. AI-assisted laboratory processes could
increase the efficiency of experiments, and support researchers to
develop and test new hypotheses.

Achieving this potential will require advances in the science of AI, the
design of AI systems that serve scientific goals, and the engineering of
such systems to operate safely and effectively in practice. These
advances in turn rely on interdisciplinary collaborations that connect
domain expertise to the development of machine learning models, and feed
the insights generated by such models back into the domain of study. As
interest in the potential of AI to drive a new wave of research grows,
the challenge for the field is to identify technical and operational
strategies to realise this potential. In the process, new questions
arise about the future of `AI for science'; whether this will emerge as
a distinct field, characterised by its own research agenda and
priorities, or whether its benefits can be best achieved through
separate, domain-focused sub-fields, which seek to integrate AI into
business-as-usual across research disciplines.

In response, this document proposes a roadmap for `AI for science'.
Synthesising insights from recent attempts to deploy AI for scientific
discovery, it proposes a research agenda that can help develop more
powerful AI tools and the areas for action that can provide an enabling
environment for their deployment. It starts by exploring core research
themes -- in simulation, causality, and encoding domain knowledge --
then draws from these ideas to propose a research agenda and action plan
to support further progress. The ideas presented are inspired by
discussions at `Machine Learning for Science: Bridging Mechanistic and
Data Driven Modelling Approaches', a Dagstuhl seminar convened in
September 2022 (see Annex 1). Abstracts from the talks given at the
seminar are shown throughout this document. These talks and the
discussions they provoked should be credited for the ideas that have
shaped it. Thank you to the speakers and participants for their
thoughtful contributions to both the seminar and the development of this
work.

\section{Snapshots of AI in science}\label{snapshots-of-ai-in-science}

Across domains, AI is being deployed to advance the frontiers of
science. The snapshots below introduce some current areas of research in
AI for science, and explore the issues raised by these research
projects. Across these snapshots, some common themes emerge:

\begin{itemize}
\item
  How can researchers most effectively combine observations, data-driven
  models, and physical models to enhance understanding of complex
  systems? To answer this question, methods are needed to integrate
  different types of model, operating across different levels of
  granularity, while managing the impact of the uncertainties that
  emerge when a machine learning model is integrated in a wider system.
  New approaches to simulation and emulation can support progress in
  tackling these challenges, alongside new strategies for examining the
  robustness or performance of machine learning models.
\item
  How do the outputs from an AI system align with what researchers
  already know about the world, and how can such systems help uncover
  causal relationships in data? Advances in causal machine learning are
  needed to connect the laws and principles already established in many
  areas of research with data-driven methods.
\item
  How can AI be integrated into the scientific process safely and
  robustly? Effective integration will rely on the ability to encode
  domain knowledge in AI systems, the design of interfaces that
  facilitate interaction between humans and AI, and the development of
  mechanisms for sharing knowledge and know-how about how to use AI in
  practice.
\end{itemize}

\subsection{In Earth sciences}\label{in-earth-sciences}

\noindent\textbf{The Earth is a complex system},\footnote{This example is
  inspired by Markus Reichstein's talk, the abstract for which is
  provided later in this document.} comprised of terrestrial, marine,
and atmospheric biospheres that interact with each other and are shaped
by biological, chemical, and physical processes that exchange energy
across scales from the molecular to the planetary. It is also a unique
system: researchers have yet to discover other planets that replicate
its dynamics. Studies of the Earth system therefore rely on observations
and physical models, which describe the dynamics of energy exchange from
first principles and use those principles to build models of the Earth's
sub-systems. As climate change perturbs this complex system, it is
increasingly important to have accurate models that can be used to
analyse how the Earth will respond to increasing carbon dioxide levels.
The challenge for Earth system science is to build more complex models
that represent the web of relationships between biospheres under
changing conditions, without generating overwhelming uncertainties and
while generating actionable insights that can be used by individuals,
organisations, and policymakers to understand the localised impact of
changing environmental conditions \cite{Summers-localized22}.

For example, how much carbon dioxide is absorbed by different biospheres
can be affected by diverse factors including volume and type of
vegetation cover, water and drought stress in different areas, and local
temperature, which have implications for how carbon dioxide contributes
to climate change. Researchers have access to data that describes local
uptake of carbon dioxide by some ecosystems, such as tropical
rainforest, European beech forest, or Mediterranean savanna, for
example, but lack sufficient observational coverage to scale from these
local observations to accurate global representations of carbon
exchange. One response to this challenge is to leverage data-driven
models to knit together the different mechanistic models that describe
(for example) carbon, water, and energy cycles in different biospheres.

By starting with observational data and combining this with
physics-informed modelling, researchers can leverage machine learning to
create simulations that can generate new understandings of how complex
systems function. Taking this approach, the FLUXNET project combines
observed data on carbon emissions from different sources to generate a
data-driven picture of global carbon dynamics. By combining data across
scales to establish a statistical model of global carbon dynamics, this
project can generate simulations of how the Earth breathes \cite{Beer-terrestrial10}. The ability to integrate across scales and
combine models of different Earth sub-systems can also contribute to
wider efforts to build a `digital twin' of the Earth, with the aim of
better understanding the implications of climate change across
biospheres and communities.

\noindent\textbf{As the Earth's climate changes},\footnote{This example is
  inspired by Markus Reichstein's talk, the abstract for which is
  provided later in this document.} researchers anticipate that local
environmental conditions will change and extreme weather events will
increase. Understanding the impact of these changes is important for
those seeking to develop appropriate responses, for example developing
environmental management plans or planning human activities.

How a landscape responds to changing environmental conditions will vary
depending on the local climate, characteristics of the terrain
(vegetation type, for example), and human activities in the area. Under
changing climate conditions, as extrapolation beyond known limits
becomes necessary, the assumptions or abstractions that form the basis
of a model can be rendered invalid. Relying solely on either mechanistic
descriptions of the system -- the impact of temperature on plant growth,
for example\footnote{Under conditions of extreme temperature, patterns
  of stomatal opening and closing in plants changes. See, for example \cite{Marchin-extreme22}.} -- or statistical models could
result in inaccuracies. Machine learning can help respond to this
challenge, through the creation of hybrid models that combine an
understanding of the physical laws with model parameters learned from
data. Researchers often already have access to known physical parameters
for a system (for example, the equations that govern how water
evaporates to air). These parameters can be fed into a machine learning
model that will learn other patterns. Known equations specify the
chemical and physical processes; machine learning can then help
elucidate the other biological forces at play. Integrating this physical
structure in the model helps make it both more interpretable to the
domain scientists and more reliable in its predictions. The resulting
model can accurately forecast the impact of climate change on the
features of local landscapes, operating within the bounds set by the
laws of physics \cite{Requena-predicting19}.

\noindent\textbf{Ice loss}\footnote{This example is inspired by Ieva
Kazlauskaitė's talk, the abstract for which is provided later in this
document.} has been the greatest contributor to sea-level rise in
recent decades \cite{Portner-ipcc19}. Large volumes of fresh water
are stored as ice: NASA estimates that if all the world's glaciers and
ice sheets melted, sea levels globally would rise by over 60 metres,
flooding all coastal cities \cite{NASA-understanding}. Researchers
can estimate the contribution that melting ice makes to sea level rise
through mechanistic models that describe the underlying physical
processes (that turn ice to water) and through observational data
about the velocity of ice sheet movement. Machine learning could offer
a toolkit to make these models more accurate, connecting ice sheet
models to ocean and atmospheric models, and integrating different data
types in hybrid mechanistic-data models.

Efforts to build such models, however, illustrate the complexity of
designing tools to meet domain needs. Projects in this space have
considered emulating the ice sheet system -- or its individual
components -- to see if models could be run faster; though successful
methodologically, it has not been clear that such efforts address a
clear research need. Another approach is to use machine learning to
streamline simulations, for instance by identifying the most effective
level of granularity for different models (is a spatial breakdown of 5km
or 10km more interesting?). An important lesson from such collaborations
is the specificity of domain needs: machine learning is a tool for
research, but just because researchers have a hammer, does not mean
every research problem is a nail. Effectively deploying machine learning
for research requires both suitable AI toolkits and an understanding of
which toolkits are best deployed for which challenges.

\subsection{In environmental and agricultural
sciences}\label{in-environmental-and-agricultural-sciences}

\noindent\textbf{Poultry farming}\footnote{This example is inspired by Dina
  Machuve's talk, the abstract for which is provided later in this
  document.} is a vital source of income and food for many communities
in Tanzania. 4.6 million households in the country raise approximately
36 million chickens, but despite the importance of this activity,
poultry farming suffers from relatively low productivity due to the
prevalence of disease. Efforts to tackle poultry diseases such as
Salmonella, Newcastle disease, and coccidiosis are held back by the
accessibility of diagnostic processes and lack of data. Diagnosis
currently requires lab analysis of droppings, which can take 3-4 days.
Once disease is confirmed, farmers often lose their entire farm's flock.

Farm-level tests and diagnostics could increase the effectiveness of
disease surveillance and treatment, giving farmers rapid access to
information about the diseases affecting their flock and action plans
about how to manage outbreaks. With mobile phones ubiquitous across the
country -- there are almost 49 million mobile phone subscriptions in
Tanzania -- there are opportunities for new uses of local data to detect
disease outbreaks.

By collecting images of droppings from farms, researchers have been
creating a dataset to train a machine learning system that can
identify the symptoms of these diseases. Fecal images are taken on
farms, annotated with diagnostic information from agricultural disease
experts and the results of lab tests, then used to train an image
recognition system to automate the diagnosis process
\cite{Machuve-poultry22}. System robustness and accuracy is vital,
given the significant implications of a positive diagnosis, and
careful design is necessary to incentivise farmers to make use of the
app.

Collaboration with experts from different domains is central to
developing this system. Input from farmers is needed to collect data and
test the system in practice; from veterinary pathologists to help
annotate the data and ensure the system's accuracy; and from
technologists to develop an AI system that is effective in deployment as
an app on mobile phones. These collaborations also open opportunities
for new forms of citizen science, as farmers and local communities are
engaged in efforts to develop and maintain an open toolkit for disease
diagnosis, providing a gateway for communities to take ownership of
machine learning as a tool to serve their needs.

\noindent\textbf{Trees and forests}\footnote{This example is inspired by
  Christian Igel's talk, the abstract for which is provided later in
  this document.} play a crucial role in maintaining healthy ecosystems.
Despite this, an estimated ten million hectares of forest are lost
globally each year due to reforestation, with only around half of this
balanced by tree-planting efforts \cite{Ritchie-forests21}. Africa
experienced an annual rate of forest loss of approximately 3.9 million
hectares per year from 2010-2020. This loss has implications for
biodiversity and people, with trees a vital contributor to ecosystem
services such as carbon storage, food provision, and shelter. In this
shifting landscape, understanding the number and distribution of trees
is important for the development of forestry management plans and for
understanding the carbon storage implications of changes to land use.

To estimate the number and biomass of trees in the West African Sahara
and Sahel, researchers have used satellite imagery of 90,000 trees from
400 sampling sites to create a labelled dataset for use in machine
learning. Using an image segmentation tool to identify the location of
trees, an automated system was able to count the number of trees, with
domain experts guiding the system to distinguish trees from surrounding
vegetation. This tree count can then be used to estimate the biomass of
trees in the area, and predict the amount of carbon they store; the
prediction is generated using allometric calculations, which translate
the properties of the tree to its carbon storage potential. In this
approach, machine learning measures the properties of the ecosystem from
satellite images, then these properties are used to feed mechanistic
models that describe the ecosystem's physical functions \cite{Brandt-trees20}. This opens the
possibility of new tools to estimate tree cover, leveraging these
insights for more effective environmental management. However, in the
process, care is needed to manage the type and nature of the
uncertainties created by different modelling approaches. Different
allometric models, for example, can be more or less suited to different
types of tree cover \cite{Hiernaux-allometric23},
meaning that the method for estimating biomass from satellite imagery
can be subject to biases when applied across a large area. A small error
in the calculation of the biomass from one tree can have a cumulatively
large effect when that method is scaled to country-level. The type and
nature of such uncertainties need to be considered when a machine
learning model is used within a wider system.

\noindent\textbf{Vector borne diseases}\footnote{This example is inspired by
  Christian Igel's talk, the abstract for which is provided later in
  this document.} account for more than 17\% of diseases in people and
over 700,000 deaths annually \cite{Hernandez-taking22}.
Changes to the climate and patterns of land use, amongst other factors,
are bringing human populations into contact with new vectors of disease.
In Africa, for example, populations of mosquitoes carrying malaria that
might previously have been found mainly in rural areas are spreading
into cities.

Tools to characterise building features from satellite imagery have
already been developed and made available for use.\footnote{For example: \cite{Quinn-mapping21}.} Leveraging these to analyse multi-scale data -- from household to
city-level---researchers are investigating how the built environment
influences people's risk of contracting mosquito-borne disease. For
example, it has been found that the prevalence of mosquitos in an area
is related to the type of roofing used in construction; metal roofing
tends to be associated with lower mosquito prevalence, potentially due
to the high temperatures they attract during the day \cite{Lindsay-reduced19}.
These insights can be deployed by policymakers in the development of
appropriate policy responses \cite{RDA-combat22}.

Decisions made on the basis of insights generated by machine learning
models will be influenced by the assumptions made in those models. In
the context of housing, for example, the decision about which type of
housing to identify as `at risk' or which building materials to flag as
`problematic' may have significant consequences for individuals or
communities. When those decisions are assimilated within a model or
analysis before a downstream `policy decision', the implications for
those communities of different courses of action may be obscured,
creating a risk of marginalising or disadvantaging individuals or
groups. The assumptions are built into the model, and how visible those
assumptions are made to different user groups, can have significant
social and scientific consequences.

\subsection{In physical sciences}\label{in-physical-sciences}

\noindent\textbf{Understanding the nature of dark matter}\footnote{This example
  is inspired by Siddharth Mishra Sharma's talk, as well as insights
  from Gilles Louppe's talk, the abstracts for which are provided later
  in this document.} is one of the biggest unsolved challenges of
particle physics today. The matter that researchers can measure using
cosmological observations makes up about 5\% of the Universe \cite{NASA-dark}.
While not directly observable, evidence for the existence of dark matter
can be found in a variety of phenomena not otherwise accounted for by
currently known laws of physics: stars rotate around galaxies faster
than might be expected; the pattern of fluctuations in primordial
microwave observations indicate that there were sources of gravitation
in the early Universe beyond ordinary matter; light bends around galaxy
clusters due to gravitational effects from dark matter.

Despite knowing that dark matter exists and that it plays an important
role in how the Universe formed, its particle composition or properties
remains unclear. Investigating these properties is the focus of
large-scale experimental studies, for example in particle
colliders.\footnote{For example: \cite{ATLAS-experiment08}} A variety of data could
contain information about the properties of dark matter, from studies of
cosmic rays, cosmic microwave radiation, properties of stars,
gravitational lensing studies, and more. These datasets are complex:
they are typically high-dimensional, represent complex relationships
between the micro-physics and macro-phenomenon in a system, and may
contain artefacts or noise from the instruments used to collect them. To
make use of this data, researchers need to account for this complexity
and tether their models to assumptions about physical processes.

The challenge for machine learning in astro-particle physics research is
to extract insights about the particle composition of dark matter from
the macroscopic patterns that can be observed in the Universe. For
example, gravitational lensing is a phenomenon in which the pathway of
light traveling through the Universe is deflected due to the influence
of gravity from an intervening mass, distorting how this background
light is observed \cite{Mishra-strong22}. Gravitational lensing
effects arising from dark matter clumps (``substructure'') could hold
information about the structure of dark matter at a microscopic level.
To infer the presence of substructure of these lensing systems,
researchers need models that describe the effect of dark matter,
ordinary matter, and the wider environment while simultaneously
modelling the form of the background light, which can be a
morphologically-complex galaxy. By letting a machine learning model,
like a neural network, describe the complex background light source, it
is possible to make predictions about how the light might appear after
being lensed with and also without the impact of dark matter clumps. By
performing many simulations considering various possibilities,
researchers can compare these with observations from telescopes and
understand which dark matter theories are compatible with the data.

Rapid progress in this field is generating a variety of models and
approaches. In its next wave of development, further research is needed
to test how trustworthy these methods are, by assessing their
performance in generating physically plausible results and robust
constraints on the properties of dark matter and other forms of new
physics \cite{Dworkin-cosmology22}.

\noindent\textbf{How particles move}\footnote{This example is inspired by
  Francisco Vargas's talk, the abstract for which is provided later in
  this document.} across their environment is a shared area of interest
for many domains. In chemistry, for example, researchers are often
interested in how molecules diffuse, and where they end up distributed,
based on the physical forces that shape their movement over time. The
analogy of particle movement can also be applied as an abstraction of
larger scale physical processes, such as in agent-based models for crowd
simulation.\footnote{Examples of agent-based models for crowd simulation
  include: \cite{Makinoshima-crowd22,Malleson-simulating20}.} In these systems the initial system state is
represented in an initial probability distribution, the scientific
objective can then also be represented as a target distribution. The
dynamics underpinning this diffusion are formalised mathematically in
the Schrödinger bridge problem. This long-standing problem is concerned
with finding the most likely paths along which particles move from their
starting distribution to their distribution at a defined point in time,
based on experimentally-observed start and end positions. In general,
finding analytic solutions to the Schrödinger bridge problem is
intractable, but machine learning tools are providing new approaches for
finding approximate numerical solutions that can be deployed across
domains \cite{Vargas-solving21}.

\subsection{In biological sciences}\label{in-biological-sciences}

\noindent\textbf{The development and differentiation of cells into tissues and
organs}\footnote{This example is inspired by Maren Büttner's talk, the
  abstract for which is provided later in this document.} is a
complicated process, shaped by hormonal and genetic influences on cell
growth \cite{Krishnamurthy-development15}.
Advances in genomics have allowed researchers to characterise the
genetic material of different organisms; more recent progress in
single-cell genomics extends this ability to the single-cell level,
unlocking detailed analysis of how genetic activity determines cellular
function.

Single-cell RNA studies examine how ribonucleic acids (RNA) shape
cellular properties and development pathways. The RNA profiles created
by genetic sequencing techniques allow researchers to identify which
genes are active in a cell. The question for the field today is how to
move from these single-cell analyses to an atlas of cell development
that shows how cells specialise and form tissues or organs.

By combining statistical and machine learning techniques, researchers
can reconstruct the gene dynamics -- which genes are activated at which
time -- that influence cell development \cite{Haghverdi-diffusion16}.
Cells in the small intestine, for example, undergo a pattern of
differentiation that takes them from their base state to highly
specialised units, able to variously secrete mucus, absorb nutrients,
or respond to hormones. By studying what genes are expressed in a cell
at an early stage, researchers can predict how the cell will specialise
and identify which genetic changes are associated with that
specialisation, opening opportunities to treat intestinal
diseases \cite{Bottcher-noncanonical21}.

Building these models relies on effective data management. Lab processes
can inject artefacts into datasets, for example batch effects arising
from how cells were grown or harvested for study, which need to be
removed from data before analysis. Effective data correction maintains
biologically-relevant information, while removing noise from the data. A
variety of tools exist for this correction, including regression models,
dimensionality reduction, graph methods, and deep learning. For domain
researchers to be able to identify the tools that are useful for them,
benchmarking studies are vital in identifying the most effective data
integration method for their purpose \cite{Luecken-benchmarking22}.
However, there remain open questions about how best to benchmark the
performance of a system when there are complex pipelines of analysis
involved. Understanding the end-to-end nature of an analytical pipeline
can be difficult, and new approaches to assessing performance may be
needed.

\noindent\textbf{To understand how the brain works},\footnote{This example is
  inspired by Jakob Macke's talk, the abstract for which is provided
  later in this document.} neuroscientists develop mathematical models
that describe the activity of individual neurons, and how these connect
across brain networks. Models on the mechanistic level take the form of
differential equations. These models are based on experimental data,
from experiments that examine how neurons respond to different signals
or perturbations. To build a computational model from this data, it is
first necessary to find which factors influence how a neuron acts,
creating a set of parameters that determine how the model works. This
process of finding parameters is often labour-intensive, relying on
trial-and-error, which limits researchers' ability to scale models
across complex neural networks. Machine learning can help streamline
that model definition process, by predicting which models are more
likely to be compatible with data. By automatically identifying model
parameters, researchers can rapidly develop simulations of complex
structures, such as brains or nervous systems in different
animals \cite{Gonccalves-training20}.

\subsection{Talks given during this workshop session}

\abstracttitle{Machine-learning-model-data-integration for a better understanding of the Earth System}
\abstractauthor[Markus Reichstein]{Markus Reichstein (MPI für Biogeochemistry - Jena, DE)}
\license

The Earth is a complex dynamic networked system. Machine learning, i.e. derivation of computational models from data, has already made important contributions to predict and understand components of the Earth system, specifically in climate, remote sensing and environmental sciences. For instance, classifications of land cover types, prediction of land-atmosphere and ocean-atmosphere exchange, or detection of extreme events have greatly benefited from these approaches. Such data-driven information has already changed how Earth system models are evaluated and further developed. However, many studies have not yet sufficiently addressed and exploited dynamic aspects of systems, such as memory effects for prediction and effects of spatial context, e.g. for classification and change detection. In particular new developments in deep learning offer great potential to overcome these limitations. Yet, a key challenge and opportunity is to integrate (physical-biological) system modelling approaches with machine learning into hybrid modelling approaches, which combines physical consistency and machine learning versatility. A couple of examples are given with focus on the terrestrial biosphere, where the combination of system-based and machine-learning-based modelling helps our understanding of aspects of the Earth system.

\abstracttitle{Poultry Diseases Diagnostics Models using Deep Learning}
\abstractauthor[Dina Machuve]{Dina Machuve (DevData Analytics - A, TZ)}
\license

Coccidiosis, Salmonella, and Newcastle are the common poultry diseases that curtail poultry production if they are not detected early. In Tanzania, these diseases are not detected early due to limited access to agricultural support services by poultry farmers. Deep learning techniques have the potential for early diagnosis of these poultry diseases. In this study, a deep Convolutional Neural Network (CNN) model was developed to diagnose poultry diseases by classifying healthy and unhealthy fecal images. Unhealthy fecal images may be symptomatic of Coccidiosis, Salmonella, and Newcastle diseases. We collected 1,255 laboratory-labeled fecal images and fecal samples used in Polymerase Chain Reaction diagnostics to annotate the laboratory-labeled fecal images. We took 6,812 poultry fecal photos using an Open Data Kit. Agricultural support experts annotated the farm-labeled fecal images. Then we used a baseline CNN model, VGG16, InceptionV3, MobileNetV2, and Xception models. We trained models using farm and laboratory-labeled fecal images and then fine-tuned them. The test set used farm-labeled images. The test accuracies results without fine-tuning were 83.06\% for the baseline CNN, 85.85\% for VGG16, 94.79\% for InceptionV3, 87.46\% for MobileNetV2, and 88.27\% for Xception. Finetuning while freezing the batch normalization layer improved model accuracies, resulting in 95.01\% for VGG16, 95.45\% for InceptionV3, 98.02\% for MobileNetV2, and 98.24\% for Xception, with F1 scores for all classifiers above 75\% in all four classes. Given the lighter weight of the trained MobileNetV2 and its better ability to generalize, we recommend deploying this model for the early detection of poultry diseases at the farm level. There are open questions about the deployment of the model at the farm level and potential areas for further research.

\abstracttitle{Simulation-based approaches to astrophysics dark matter searches}
\abstractauthor[Siddharth Mishra-Sharma]{Siddharth Mishra-Sharma (MIT - Cambridge, US)}
\license

We are at the dawn of a data-rich era in astrophysics and cosmology, with the capacity to extract useful scientific insights often limited by our ability to efficiently model complex processes that give rise to the data rather than the volume and nature of observations itself. I will describe recent progress in applying mechanistic forward modeling techniques to a range of astrophysical observations with the goal of searching for signatures of new physics, in particular the nature of dark matter. These leverage developments in machine learning-aided inference, e.g. using simulation-based inference as well as differentiable probabilistic programming, while encoding domain knowledge, in order to maximize the scientific output of current as well as future experiments.

\abstracttitle{Single-cell transcriptomics}
\abstractauthor[Maren B\"uttner]{Maren B\"uttner (Helmholtz Zentrum München \& Universität Bonn)}
\license

Cells are the fundamental units of life. Understanding cellular processes is a basis for improving human health, disease diagnosis and monitoring. The advent of single-cell transcriptomics (scRNA-seq) allows characterizing the gene expression patterns of entire organs and organisms at single cell resolution. The human genome encodes more than 30.000 genes, and high-throughput scRNA-seq methods create samples with tens of thousands of cell measurements. The analysis of such data requires a variety of methods from the machine learning field, e.g. dimensionality reduction techniques from PCA to variational autoencoders, graph-based clustering, classification of cell types, trajectory inference and causal inference of gene regulation to understand cell fate decision making. To date, scRNA-seq is a widely applied research technique, which has the potential for standard application in the clinics.  My presentation focusses on current approaches for large-scale scRNA-seq data, current open questions, and implications for human health.

\abstracttitle{Estimating ecosystem properties: Combining machine learning and mechanistic models}
\abstractauthor[Christian Igel]{Christian Igel (University of Copenhagen, DK)}
\license
\jointwork{Martin Brandt, Rasmus Fensholt, Compton J. Tucker, Ankit Kariryaa, Kjeld Rasmussen, Christin Abel, Jennifer Small, Jerome Chave, Laura Vang Rasmussen, Pierre Hiernaux, Abdoul Aziz Diouf, Laurent Kergoat, Ole Mertz, Fabian Gieseke, Sizhuo Li, Katherine Melo}
\abstractref[https://doi.org/10.1038/s41586-020-2824-5]{Brandt, M., Tucker, C.J., Kariryaa, A. et al. An unexpectedly large count of trees in the West African Sahara and Sahel. Nature 587, 78–82 (2020).}
\abstractrefurl{https://doi.org/10.1038/s41586-020-2824-5}

Progress in remote sensing technology and machine learning algorithms enables scaling up the monitoring of ecosystems. This leads to new knowledge about their status and dynamics, which will be helpful in land degradation assessment (e.g., deforestation), in mitigating poverty (e.g., food security, agroforestry, wood products), and in managing climate change (e.g., carbon sequestration).

We apply deep learning for the mapping of individual trees and forests. Tree crowns are segmented in satellite imagery using fully convolutional neural networks. This provides detailed measurements of the canopy area and of the distribution of trees within and outside forests. Allometric equations are applied to estimate the biomasses (and thereby the stored carbon) of the individual trees. We use iterative gradient-based optimization of the allometric models and suggest techniques such as  jackknife+ for quantifying the uncertainty of the model predictions. Tree biomass can also be directly inferred from LiDAR (laser imaging, detection, and ranging) measurements using 3D point cloud neural networks. This leads to highly accurate results without requiring a digital elevation model. 

In a new project, we consider risk assessment of vector-borne diseases based on deep learning and remote sensing. Malaria risk is related to the housing conditions, for example, the type of roofing material, which can be determined from satellite images.

\abstracttitle{Partial differential equations and Variational Bayes}
\abstractauthor[Ieva Kazlauskaite]{Ieva Kazlauskaite (University of Cambridge, GB)}
\license

Inverse problems involving partial differential equations (PDEs) are widely used in science and engineering. Although such problems are generally ill-posed, different regularisation approaches have been developed to ameliorate this problem. Among them is the Bayesian formulation, where a prior probability measure is placed on the quantity of interest. The resulting posterior probability measure is usually analytically intractable. The Markov Chain Monte Carlo (MCMC) method has been the go-to method for sampling from those posterior measures. MCMC is computationally infeasible for large-scale problems that arise in engineering practice. Lately, Variational Bayes (VB) has been recognised as a more computationally tractable method for Bayesian inference, approximating a Bayesian posterior distribution with a simpler trial distribution by solving an optimisation problem. The talk covered some recent experiences of applying Bayesian inference, generative models and probabilistic programming languages in the context of learning material properties in civil engineering and in ice sheet and ice core modelling. The main shortcomings of PPLs and differentiable problems were highlighted.

\abstracttitle{The Schrödinger bridge problem}
\abstractauthor[Francisco Vargas]{Francisco Vargas (University of Cambridge, GB)}
\license

Recent works in diffusion-based models have been achieving competitive results across generative modelling and inference, in this presentation we propose to explore a unifying framework based on Schrodinger bridges to explore/explain diffusion-based methodology. The Schrödinger bridge problem (SBP) finds the most likely stochastic evolution between two probability distributions given a prior (reference) stochastic evolution. Recently SBP based methodology has made its way into generative modelling , sampling, and inference. In this talk we propose the exploration of a unifying framework for the aforementioned works based on the renowned IPF/Sinkhorn algorithm. The motivation behind this is to cast a unifying lens via the Schrodinger perspective relating inference, sampling and transport, in a way that we can leverage many of the useful techniques and heuristics from each field to benefit each other.

\section{Building effective
simulations}\label{building-effective-simulations}

\subsection{Moving upstream}\label{moving-upstream}

Science proceeds through hypothesis, observation and analysis. For
hundreds of years, researchers have advanced the frontiers of knowledge
by collecting data, compressing those observations into a model, then
computing that model to create representations of how the world works,
generating new insights about natural and physical phenomena and
theories about the systems from which those phenomena emerge in the
process \cite{Blei-build2014}.
These mathematical models rely on numerical methods: algorithms that
help solve mathematical problems where no analytical solution is
available. Today, data collection and the basic computational tasks
involved in its analysis -- linear algebra, optimisation, simulation,
and so on -- remain consistent features of the scientific process.
Progress in machine learning, however, has changed the modelling
landscape.

`AI for science' offers a data-centric approach to modelling and
simulating the world. Operating alongside the traditional mathematical
models that are central to many disciplines, machine learning provides
data-centric analytical methods that can be integrated across the
scientific pipeline, for example enabling sophisticated simulations of
real-world systems. These simulations can be used to inform model
development, test hypotheses and shape areas of research focus, or
unlock insights from complex data.

\subsection{Nurturing a diversity of
approaches}\label{nurturing-a-diversity-of-approaches}

Simulations are a well-established tool for scientific discovery. Their
fundamental task is to allow data sampling from a model where the
differences between simulation and the real world are reduced as far as
feasible, to enable experimentation or testing of the impact of
different perturbations, while allowing some measure of simplification
of the system. Effective simulators allow researchers to move from
theory to an understanding of what data should look like.

Domains such as particle physics, protein folding, climate science, and
others, have developed complex simulations that use known theories and
parameters of interest to make predictions about the system of study. AI
for science can be brought in to speed up some of these through
surrogate models. Machine learning can complement `traditional'
approaches to scientific simulation, adding components that model the
most uncertain elements of a system to strongly mechanistic models that
might otherwise be too restrictive in their assumptions.

Much early excitement surrounding AI for science was rooted in the
reverse process, asking: instead of starting with theory, could
researchers instead start with the large amounts of data available in
many areas of research and, from that data, build an understanding of
what an underpinning theory might be? Given a set of observations, is it
possible to find parameters for a model that result in simulations that
reflect the measured data? Such simulation-based inference (SBI) offers
the opportunity to generate novel insights across scientific
disciplines.

To enable such analysis, machine learning methods are needed that can
extract insights from high-dimensional, multi-modal data, in ways that
are labour- and compute-efficient \cite{Cranmer-frontier2020}. The field of
probabilistic numerics offers a way to flexibly combine information from
mechanistic models with insights from data, solving numerical problems
through statistical approaches \cite{Hennig-probabilistic2022}. Operationalising these methods
to create effective data-driven simulations requires balancing different
model characteristics. The model's parameters must be specified to a
sufficient level of granularity to describe the real-world system, while
operating at a level of abstraction that is amenable to analysis and
computation; almost all models are `wrong' or falsifiable because of
this, but some level of abstraction is necessary to make them useful for
analysis. The simulation must also be designed to be robust, and able to
generate inferences that align with real-world observations.

\subsection{Truth, truthiness, and interfacing with the real
world}\label{truth-truthiness-and-interfacing-with-the-real-world}

The excitement underpinning AI for science stems from the aspiration to
unearth new understandings of the world, leveraging data to advance the
frontiers of knowledge. While subject to their own limitations, the
scientific community has developed checks and balances to scrutinise new
knowledge and maintain the rigour of scientific inquiry. Recent years
have seen a variety of challenges or benchmarks emerge in the machine
learning community that have come to represent the field's expected
standards of performance from algorithms on defined tasks. However,
these standards do not necessarily align with the expectations of domain
researchers \cite{Hermans-averting21}.
As data-centric simulations are integrated into scientific process,
machine learning researchers must consider their responsibility in
maintaining the integrity of the domains into which they are deployed,
raising the question: what guardrails are needed to ensure researchers
can be confident in the outputs from machine learning-enabled
simulations?

A variety of diagnostic tests can help. Core to many of these
diagnostics is analysis of whether a model is computationally faithful.
In short: the inferences generated by a simulation should reflect those
from observations \cite{Hermans-averting21}. One approach to checking this
alignment is to consider the consistency of distributions from inferred
and observed datasets. If the model is a good fit, the data it generates
should broadly match the data observed through experimentation.

Underpinning these diagnostics is a fundamental question about how to
manage uncertainty, in a context where different failure modes have
different implications. Put simply: when a model fails, is it worse to
be over-confident in its results, or over-conservative? In the
scientific context, over-confidence seems more likely to result in
negative outcomes, whether through giving misleading interpretations or
results or driving lines of enquiry in unproductive directions. Machine
learning methods can be designed for conservatism, reducing the risk of
false positives.

Implementing a schedule of model building, computing, critiquing, and
repeating can refine this process. One lesson from experiences of
building machine learning-enabled simulations is that there can be a
disconnect between how machine learning approaches inference and model
building, and how the same task is approached by domain scientists. From
a domain perspective, model building seems naturally an iterative
process: collect data, fit a model, find errors or areas for
improvement, update the model, and so on. This iterative process is
guided by expert intuition and knowledge; deep understanding of the
system under study and how it responds to perturbation. Machine learning
research has developed practices for prior elicitation --- using domain
knowledge to shape the structure of probabilistic models --- but the
nuances of this domain intuition are often not easily captured a priori,
instead emerging when models fail as an informal sense of what `feels'
like it should be true. This qualitative input is vital in building
effective simulations. It requires close collaboration, which in turn
requires an investment of time and energy from domain communities,
generated through mutual trust, incentives, and long-term
relationship-building.

\subsection{Connecting simulation to
practice}\label{connecting-simulation-to-practice}

Computational tools are central to the effective deployment of machine
learning-enabled simulation. The function and form of such tools must
align with the requirements of the community deploying them. Designing
computational systems to match user needs -- and work effectively in
practice -- requires both effective software engineering and close
collaboration with domain groups that can articulate the requirements
and expectations of those working in the field. To remain effective over
the longer-term, such systems must leverage effective software
engineering practices, including embedding version control and building
interfaces that work with other models and systems. Those practices, and
the software systems that emerge from them, must be designed for the
needs of those using the system, drawing from existing best practices in
software engineering, but adapting those practices to reflect the needs
of the domain for deployment.

Constructing computational tools requires a mix of technical insight and
craft skill -- of knowledge and know-how. Tools produced by the machine
learning community differ in their usefulness on different problems:
some work well for certain tasks, but not for others. Without access to
such craft skills, those outside the `AI for science' community can find
it challenging to determine which tools to use for which purposes,
reducing the generalisability of existing methods and approaches. This
challenge becomes particularly visible when practitioners are tightly
integrated into the analysis pipeline, such as in applications in
developmental biology, in the developing world, and in data-centric
engineering. Widening access to the field will require user guides that
characterise which simulations are effective for which tasks or
purposes, supported by case studies or user stories that help demystify
how machine learning can work in practice.

\subsection{Directions}\label{directions}

Machine learning typically requires an explicit representation of a
likelihood, but these are often difficult to compute. Further advances
in SBI are necessary to allow researchers to identify model parameters
from data.

\begin{itemize}
\item
  Techniques such as likelihood-free inference can enhance existing
  Bayesian methods for inferring posterior estimations \cite{Alsing-fast19}.
\item
  Building surrogate models,\footnote{See above, and \cite{Lavin-simulation21}.} using Bayesian
  approaches for simulation planning to optimise information
  gain,\footnote{See, for example, \cite{Cranmer-active17}.}
  or deploying emulations \cite{Boelts-flexible22}
  can also enhance the efficiency of simulations.
\item
  Probabilistic numerics offers a route to develop statistically-optimal
  algorithms that are amenable to comprehensive uncertainty
  quantification, leveraging Gaussian Process-based Ordinary
  Differential Equation (ODE) solvers to pursue simulation as an
  inference problem \cite{Kersting-uncertainty21}.
\end{itemize}

Operationalising these approaches will also require new toolkits to
support implementation of probabilistic numerical methods.\footnote{See,
  for example, the previous Dagstuhl meeting on this topic:
  \url{https://www.probabilistic-numerics.org/meetings/2021_Dagstuhl/}
  and \cite{Schmidt-probabilistic21}.}

Computational faithfulness -- alignment of inferred parameters with
scientific knowledge -- can be achieved through:

\begin{itemize}
\item
  Diagnostic checks in the self-consistency of the Bayesian joint
  distribution, which measure the scientific quality of the regions
  computed by Bayesian SBI methods \cite{Hermans-averting21,Mishra-inferring22}.
  Checking for self-consistency gives a sense whether the model is `good
  enough' (ie whether the inference engine gives a good sense of the
  posterior).
\item
  Enforcing conservative neural ratio estimation through binary
  classifier specification, producing more conservative posterior
  approximations \cite{Delaunoy-towards22}.
\item
  Hybrid modelling, which combines machine learning components learned
  from data with the mechanistic components specified by existing domain
  knowledge \cite{Wehenkel-robust22}.
\item
  Further study of the impact of model misspecification could also help
  generate new robustness diagnostic checks \cite{Cannon-investigating22}.
\end{itemize}

`Digital twins' have recently received much attention as a tool to
exploit sophisticated simulations. In Earth sciences, for example,
ambitious efforts to develop a digital twin of the Earth propose to
allow more accurate forecasting, visualisation, or scenario-testing of
the impact of climate change and efforts to mitigate it.\footnote{For
  example: \cite{European-destination22}.}
The challenge is to integrate different models or components of a system
-- for example, connecting atmospheric models, with land models, with
models of human behaviour -- in a way that represents the complete Earth
system. That requires consideration of the different levels of
granularity with which these different models operate: economic models
of human behaviour, for example, operate with different assumptions and
levels of enquiry in comparison to physical models of ocean circulation.
The full range of granularities becomes apparent when considering that
specific applications, such as disease monitoring on poultry farms, sit
within the wider ecosystem of the natural and built environment. A
digital twin needs to make choices about what levels of granularity it
is operating at, from the scale of the poultry farm to the planet. The
questions that emerge from such ambitions is: what level of granularity
is helpful or necessary to deliver effective results? And what
interfaces between diverse models might be possible?

\subsection{Talks given during this workshop session}

\abstracttitle{Information from data and compute in scientific inference}
\abstractauthor[Philipp Hennig]{Philipp Hennig (Universität Tübingen, DE)}
\license

Simulations are central to scientific inference. Simulators are typically treated as black boxes, with the inference loop wrapped around them. This approach is convenient for the programming scientists, but can be highly inefficient. Probabilistic numerical methods represent computational and empirical data in the same language, which allows for inference from mechanistic knowledge and empirical data in one combined step. I will argue that scientific computing needs to embrace such new computational paradigms to truly leverage ML in science, which also requires rethinking scientific codebases.

\abstracttitle{ODE filters and smoothers: probabilistic numerics for mechanistic modelling}
\abstractauthor[Hans Kersting]{Hans Kersting (INRIA - Paris, FR)}
\license

Probabilistic numerics (PN) unifies statistical and numerical approximations by formulating them in the same language of statistical (Bayesian) inference. For ODEs, a well-established probabilistic numerical method is ODE filters and smoothers which can help to deal more aptly with uncertainty in mechanistic modeling. In the first half of this talk, we will first introduce PN and then present ODE filters/smoothers as a specific instance of PN. In the second half, we will discuss how ODE filters/smoothers can improve mechanistic modeling in the natural sciences and present a recent application of inferring the parameters of real-word dynamical system.

\abstracttitle{Four short stories on simulation-based inference}
\abstractauthor[Jakob Macke]{Jakob Macke (Universität Tübingen, DE)}
\license

Many fields of science make extensive use of simulations expressing mechanistic forward models, requiring the use of simulation-based inference methods. I will share experiences and lessons learned from four applications: Describing the dynamics and energy consumptions of neural networks in the stomatogastric ganglion; inferring parameters of gravitational wave models; optimising single-molecule localisation microscopy, and building computational models of the fly visual system. I will try to convey some thoughts on the challenges and shortcomings of current approaches.

\abstracttitle{Towards reliable simulation-based inference and beyond}
\abstractauthor[Gilles Louppe]{Gilles Louppe (University of Liège, BE)}
\license

Modern approaches for simulation-based inference build upon deep learning surrogates to enable approximate Bayesian inference with computer simulators. In practice, the estimated posteriors' computational faithfulness is, however, rarely guaranteed. For example, Hermans et al., 2021 have shown that current simulation-based inference algorithms can produce posteriors that are overconfident, hence risking false inferences. In this talk, we will review the main inference algorithms and present Balanced Neural Ratio Estimation (BNRE), a variation of the NRE algorithm designed to produce posterior approximations that tend to be more conservative, hence improving their reliability.

\abstracttitle{Modeling the data collection process: My journey}
\abstractauthor[Thomas G. Dietterich]{Thomas G. Dietterich (Oregon State University - Corvallis, US)}
\license

In this talk, I will describe three examples of my attempts to integrate subject-matter knowledge with machine learning. The first example involves predicting grasshopper infestations. I will sketch the methodology in which we first modeled the life cycle of the grasshoppers to capture the factors that affect their population. Unfortunately, most variables of interest were not measured, so we used the model to guide the construction of proxy variables. Ultimately, this project did not succeed, but it is hard to determine whether this is due to modeling problems or to the chaotic nature of the biological phenomenon.

\section{Connecting data to causality}\label{connecting-data-to-causality}

\subsection{Causality in science and data}\label{causality-in-science-and-data}

Most scientific endeavours have a causal element: researchers want to
characterise how a system works, why it works that way, and what happens
when it is perturbed. How researchers identify cause-and-effect
relationships varies across domains. For some disciplines, the process
of hypothesis design -- data collection -- model development provides
the core structure for interrogating how a system works. In others,
where experimentation is more difficult, researchers may rely on natural
experiments and observations to compare the response of a system under
different conditions. Those studying the Earth system, for example, have
little scope to replicate planetary conditions, so instead rely on
observational data and modelling to identify the impact of different
interventions. These different approaches, however, share a modelling
approach in which researchers provide variables to create structural,
causal models.

In contrast, machine learning proceeds by learning representations or
rules from data, based on statistical information, rather than
structured rules about how a system works (such as physical laws).
Causal inference -- the ability to identify cause-and-effect
relationships in data -- has been a core aim of AI research, in service
of both wider ambitions to replicate intelligence in machines and
efforts to create AI systems that are robust in deployment. However, in
many respects efforts to integrate causal inference into AI systems have
yet to deliver \cite{Scholkopf-representation21}.

An apocryphal story in AI tells of efforts by US researchers during the
1980s to train a computer system that could distinguish between images
of tanks from the US and USSR. The resulting system delivered high
accuracy on its training data, but failed repeatedly in practice. The
system was subsequently found to be classifying images based on their
resolution and background features -- is the image grainy? Does it
contain snow? -- rather than the tanks themselves. It found patterns in
the data that were co-incident, rather than causal. That same error has
real-world implications for the AI systems deployed today. In medical
sciences, AI systems trained to detect collapsed lungs from medical
images have been proven inaccurate, after the model was found to have
learned to detect the tube inserted into the lung to enable a patient to
breath as a response to its collapse, rather than the physical features
of the lung itself \cite{Rueckel-impact20}.
In medical sciences, deployment of such systems could put patient care
at risk. In social sciences, these AI design and data bias failures can
combine to marginalise vulnerable populations \cite{Emspak-prejudice16}.

Conversely, an understanding of the structures within data can improve
the accuracy of machine learning analyses. In exoplanet discovery, for
example, machine learning is used as a tool to detect variations in
light signals from large-scale astronomical datasets. The movement of
exoplanets around stars results in periodic changes to the light signals
from those stars, as the planet obscures them in its transit. Machine
learning can detect those signals and predict where exoplanets might be
located, but the data is often noisy. Noticing that the structure of
this noise was consistent across a number of stars, which were too
distant from each other to be interacting, researchers concluded that
instrumentation effects were distorting the data, and developed a method
to model those effects and remove them from exoplanet predictions. The
result was an efficient method for exoplanet identification that
subsequently contributed to the discovery of the first potentially
habitable planet \cite{Scholkopf-causality22}.

\subsection{Causal models as a route to advancing the science of AI and
AI for
science}\label{causal-models-as-a-route-to-advancing-the-science-of-ai-and-ai-for-science}

Many of these errors in misdiagnosing cause-effect relationships arise
from a core assumption in many machine learning methods: that data
follows an independent and identical distribution (IID). In practice,
almost all data from real-world, or complex, systems will violate this
assumption, given the interconnectedness of different variables. The
task of causality in machine learning is to create models that can
manage this violation, distinguishing between patterns in data that
simply co-occur and patterns that are causal. The resulting AI systems
would be able to solve a task in many different environments, based on
an understanding of the fundamental causal mechanisms in a
system \cite{Peters-elements17}. They would be more robust in
deployment, being less likely to make incorrect predictions as the
environment in which they operate changes, and could be more efficient
to train and deploy. They would also represent a step towards
replicating human- or animal-like intelligence, being able to solve a
task in many different environments.

In these regards, causal machine learning offers a route to balancing
the widespread utility of statistical modelling with the strengths of
physical models. Causality allows models to operate at a level of
abstraction beyond strongly mechanistic approaches, such as those based
on differential equations, moving along a continuum from mechanistic to
data-driven modelling. They provide researchers with the ability to make
accurate predictions under conditions of dataset shift (enable out of
distribution generalisation); can provide insights into the physical
processes that drive the behaviour of a system; unlock progress towards
AI systems that `think' in the sense of acting in an imagined space;
while also leveraging insights that can be learned from data, but not
otherwise detected.\footnote{For reference, see the table on page 11 of
  reference \cite{Scholkopf-causality22}.} They also offer opportunities to explore
counterfactuals in complex systems, asking what the impact of different
interventions could have been, opening a door to the development of
simulation-based decision-making tools.\footnote{Such tools may have
  particular relevance in policy. For example: \cite{Mastakouri-causal20}.}

Achieving this potential requires technical developments in a number of
directions, but can also yield more effective AI systems. Such systems
would:

\begin{itemize}
\item
  Be able to operate on out of distribution data, performing the task
  for which they are trained in environments with varying conditions.
\item
  Be able to learn how to perform a task based on relatively few
  examples of that task in different conditions, or be able to rapidly
  adapt what they have learned for application in new environments
  through transfer, one-shot, or lifelong learning approaches.
\item
  Support users to analyse the impact of different interventions on a
  system, providing explanations or ways of attributing credit to
  different actions.
\item
  Respond to different ways of transmitting information between
  individuals and groups, enabling effective communication with their
  users or other forms of cultural learning.
\end{itemize}

\subsection{From methods to
application}\label{from-methods-to-application}

Achieving the level of technical sophistication required for causal
modelling requires careful model design, based on close collaboration
between machine learning and domain scientists. The process of specifying
what to represent in a causal machine learning system involves a series
of `micro-decisions' about how to construct the model, negotiated by
integrating machine learning and domain expertise. In this regard,
causal machine learning can be a positive catalyst for deeper
interdisciplinary collaboration; model construction can be a convening
point for sharing understandings between domains. However, the level of
detail required can also be in tension with efforts to promote
widespread adoption of AI methods across research. The availability of
easy-to-use, off-the-shelf AI tools has been an enabler for adoption in
many domains. The hand-crafted approach inherent to current causal
methods renders them less accessible to non-expert users. Part of the
challenge for the field is to make such methods more broadly accessible
through open-source toolkits or effective software engineering
practices.

This tension between specification and learning also highlights the
importance of nurturing a diversity of methods across the spectrum from
data-driven to mechanistic modelling. The domain (or, how much prior
knowledge is available and what knowledge should be included), research
question of interest, and other practical factors (including, for
example, compute budget), will shape where along this spectrum
researchers wish to target their modelling efforts.

While pursuing practical applications, advances in causal inference
could help answer broader questions about the nature of intelligence and
the role of causal representations in human understanding of how the
worlds work. Much of human understanding of the world arises from
observing cause and effect; seeing what reaction follows an intervention
-- that an object falls when dropped, for example -- in a way that
generalises across circumstances and does not require detailed
understanding of mathematical or physical laws. Integrating this ability
into machine learning would help create systems that could be deployed
on a variety of tasks. The process of building causal machine learning
forces researchers to interrogate the nature of causal representations
-- What are they? How are they constructed from the interaction between
intelligent agents and the world? By what mechanism can such agents
connect low-level observations to high-level causal variables? -- which
may in turn support wider advances in the science of AI.

\subsection{Directions}\label{directions-1}

Causality in machine learning is a long-standing and complex challenge.
In the context of scientific discovery, learning strategy, model design,
and encoding domain knowledge all play a role in helping identify
cause-effect relationships.

Different learning strategies can improve the `generalisability' of
machine learning, increasing its performance on previously unseen tasks,
based on learning underlying structure of a task or environment in ways
that can contribute to broader understandings of causality. Such
learning strategies include:

\begin{itemize}
\item
  Transfer learning, taking learning from one task or domain and
  applying it in another.
\item
  Multi-task learning, enabling a system to solve multiple tasks in
  multiple environments.
\item
  Adversarial learning, to reduce the vulnerability of models to
  performance degradation on out-of-distribution data.
\item
  Causal representation learning, defining variables that are related by
  causal models \cite{Scholkopf-causality22}.
\item
  Reinforcement learning strategies that reward agents for identifying
  policies based on invariances over different conditions.
\end{itemize}

Across these new learning approaches, attempts to establish causal
mechanisms are also prompting progress in machine learning theory,
through statistical formulations of core principles \cite{Guo-causal22}.

Combining different methods can also enhance the functionality of an AI
system. For example:

\begin{itemize}
\item
  Neural ODEs have been shown to identify causal structures in time
  series data \cite{Aliee-beyond21}.
\item
  Describing causal effects as objective functions in constrained
  optimisation problems can deliver a form of stochastic causal
  programming \cite{Padh-stochastic22}.
\item
  Technical interventions \cite{Jakobsen-distributional22}
  can constrain or optimise a model towards causal outcomes. As with
  simulation design, diagnostic checks can also help identify
  cause-effect relationships by examining model outputs against `reality
  criteria',\footnote{Including syntactic, semantic, and pragmatic
    elements: \cite{Stadler-wirklichkeitskriterien90}.} which compare outputs to
  real-world results.
\end{itemize}

There are also a variety of approaches to representing existing
scientific knowledge in machine learning models, notably by specifying
the assumptions made about the world through symmetries, invariances,
and physical laws (see Figure 1).

\subsection{Talks given during this workshop session}

\abstracttitle{Causality, causal digital twins, and their applications}
\abstractauthor[Bernhard Sch\"olkopf]{Bernhard Sch\"olkopf (MPI für Intelligente Systeme - Tübingen, DE)}
\license

\begin{enumerate}
\item Desiderata for causal machine learning: work with (and benefit from) non-IID data, multi-task/multi-environment, sample-efficient, OOD, generalisation from observation of marginals, interventional.
 \item Modelling taxonomy: differential equations, causal models, statistical models.
 \item How to get from one level to the next.
\item How to transfer between statistical models that share the same underlying causal model. 
\item The assumption of independent causal mechanisms (ICM) (for example, invariance/autonomy) and sparse mechanism design. 
 \item How to derive the arrow of time from ICM and algorithmic information theory.
 \item Statistical formulation of ICM: causal de Finetti.
 \item Application to exoplanet discovery and Covid-19 vaccine scenarios.
 \item Causal representations as (a) causal digital twins and (b) AI models.
\end{enumerate}

\abstracttitle{Invariance: From Causality to Distribution Generalization}
\abstractauthor[Jonas Peters]{Jonas Peters (University of Copenhagen, DK)}
\license

Assume that we observe data from a response $Y$ and a set of covariates $X$ under different experimental conditions (or environments). Rather than focusing on the model that is most predictive, it has been suggested to take into account the invariance of a model. This can help us to infer causal structure (Which covariates are causes of $Y$?) and find models that generalize better (How well does the model perform on an unseen environment?). We show a few applications of these general principles and discuss first steps towards understanding the corresponding theoretical guarantees and limits.

\abstracttitle{Can we discover dynamical laws from observation?}
\abstractauthor[Niki Kilbertus]{Niki Kilbertus (TU M\"unchen, DE \& Helmholtz AI M\"unchen, DE)}
\license

I will start with a brief introduction to identifiability of ODE systems from a unique continuous or discrete observed solution trajectory. Then, I will provide an overview of modern approaches to inferring dynamical laws (in the form of ODEs) from observational data with a particular focus on interpretability and symbolic methods. Finally, I will describe our recent attempts and results at inferring scalar ODEs in symbolic form from a single irregularly sampled, noisy solution trajectory.

\abstracttitle{Invariances and equivariances in machine learning}
\abstractauthor[Soledad Villar]{Soledad Villar (Johns Hopkins University - Baltimore, US)}
\license

In this talk, we give an overview of the progress in the last few years by several research groups in designing machine learning methods that repeat physical laws. Some of these frameworks make use of irreducible representations, some make use of high-order tensor objects, and some apply symmetry enforcing constraints. Our work shows that it is simple to parameterise universally approximating functions that are equivariant under actions of the Euclidean, Lorentz, and Poincare group at any dimensionality. The key observation is that $O(d)$-equivariant (and related group-equivariant) functions can be universally expressed in terms of a lightweight collection of dimensionless scalars (scalar products and scalar contractions of the scarla, vector, and tensor inputs). We complement our theory with numerical examples that show that the scalar-based method is simple and efficient, and mention ongoing work on cosmology simulations.

\abstracttitle{Divide-and-Conquer Equation Learning with R2 and Bayesian Model Evidence}
\abstractauthor[Bubacarr Bah]{Bubacarr Bah (AIMS South Africa - Cape Town, ZA)}
\license

Deep learning is a powerful method for tasks like predictions and classification, but lacks interpretability and analytic access. Instead of fitting up to millions of parameters, an intriguing alternative for a wide range of problems would be to learn the governing equations from data. Resulting models would be concise, parameters can be interpreted, the model can adjust to shifts in data, and analytic analysis allows for extra insights. Common challenges are model complexity identification, stable feature selection, expressivity, computational feasibility, and scarce data. In our work, the mentioned challenges are addressed by combining existing methods in a novel way. We choose multiple regression as a framework and argue how a surprisingly large space of model equations can be captured. For feature selection, we exploit the computationally cheap coefficient of determination (R2) to loop through millions of models, and by using a divide-and-conquer strategy, we are able to rule out remaining models in the equation class. Final model selection is achieved by exact values of the Bayesian model evidence with empirical priors, which is known to identify suitable model complexity without relying on mass data. Random polynomials, and a couple of chaotic systems are used as examples.

\section{Encoding domain knowledge}\label{encoding-domain-knowledge}

\subsection{Where's My {[}Science{]} Jetpack?}\label{wheres-my-science-jetpack}

Humans have a long history of imagining futures where human progress is
accelerated by intelligent machines. Embedded in these visions for the
future are aspirations that AI can be a faithful servant, easing daily
activities or enhancing human activities \cite{Royal-narratives18}.
As with many emerging technologies, the reality of AI today looks
different to these Sci-Fi futures.\footnote{The title of this section is
  inspired by:
  \url{https://www.fantasticfiction.com/w/daniel-h-wilson/where-s-my-jetpack.htm}}
Practical experiences of deploying AI highlights a range of potential
failure modes, often rooted in insufficient contextual awareness,
misspecification of user needs, or misunderstanding of environmental
dynamics \cite{Paleyes-challenges22}.

Today's science builds on thousands of years of attempts to understand
the world, which can be leveraged to design AI that serves scientific
goals. The result should be a collaborative endeavour between humans and
machines. Researchers need the analytical power of AI to make sense of
the world, while AI needs input from human understandings of the domain
in which it is deployed to function effectively; both need well-designed
human-machine interfaces to make this collaboration work. In this
context, effective integration of domain knowledge into AI systems is
vital, and three (broad) strategies have emerged to facilitate this
encoding: algorithmic design; AI integration in the lab; and effective
communication and collaboration.

\subsection{Encoding domain knowledge through model
design}\label{encoding-domain-knowledge-through-model-design}

Traditional modelling approaches make use of well-defined rules or
equations that explain the dynamics of the system under study. The laws
of physics, for example, describe how energy moves through a system,
based on conservation principles. These laws are complemented by
mathematical symmetries that arise from our abstract representations of
physical objects and describe what features of an object remain
consistent, despite changes or transformations in a system \cite{Villar-scalars21}.
There may also be known invariances in a system: factors that do not
change under any perturbations or that change in a defined
way \cite{Ling-machine16}.
Building on this existing knowledge, and connecting to efforts to
generate causal understandings of the world through machine learning, an
area of growing interest has been the design of machine learning models
that respect these rules or symmetries.

The principle underpinning this design strategy is that it is possible
to move across a continuum from statistical (data-driven) models to
strongly mechanistic models, creating hybrid systems whose outputs
should be constrained by what is physically feasible, while also
leveraging insights from data (Figure 1).

\begin{figure}
\begin{center}
\includegraphics[width=0.8\textwidth]{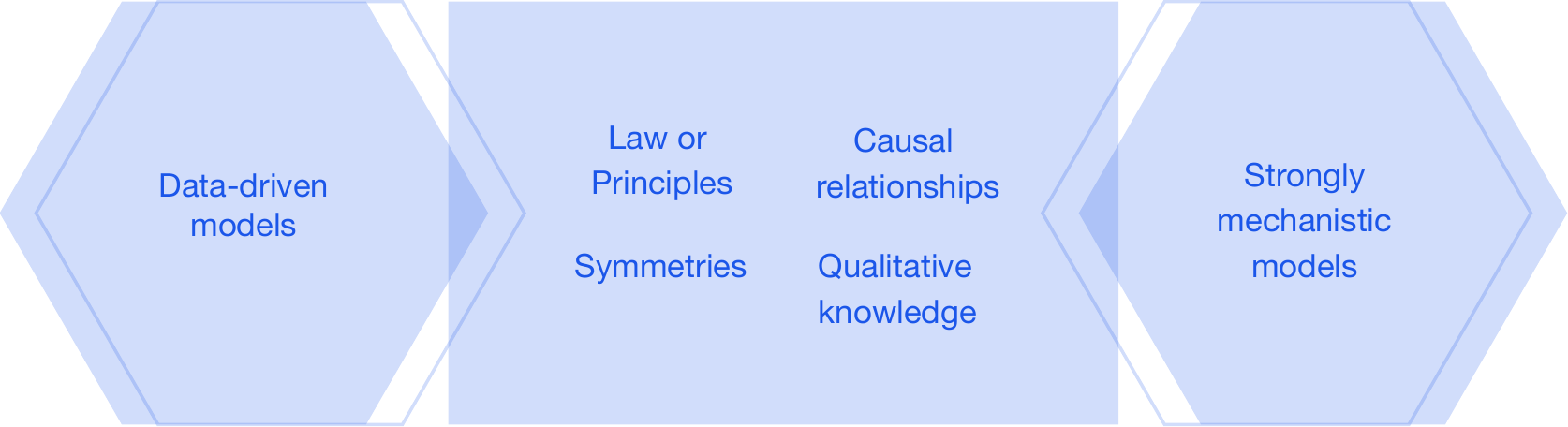}
\end{center}
\caption{Models along a spectrum from classical i.i.d models to strongly mechanistic differential equation models introduce aspects of causality and symmetries to create a continuum between mechanistic and data-driven worlds. Statistical or data-driven models are weakly mechanistic (i.e. they include smoothness assumptions or similar).}
\label{model-spectrum}
\end{figure}

At one end of that continuum, mechanistic models would obey known laws
or principles in a strongly deterministic way; at the other, statistical
models encode fewer assumptions and rely more on data \cite{Lawrence-licsbintro10}. The addition of
invariances and symmetries, alongside other forms of domain knowledge,
allows bridging between these two model classes (Figure \ref{model-spectrum}). Models that
describe how much heat is absorbed by the oceans under conditions of
climate change, for example, should obey the laws of thermodynamics and
energy conservation. By encoding the domain knowledge that has yielded
these fundamental laws, such as the conservation of momentum or energy,
researchers can ensure the outputs of a machine learning model will have
a physically allowable expression. This encoding can come from
integrating equations, symmetries, or invariances into model design.
These encodings constrain the operation of a machine learning system to
align with the known dynamics of physical systems. The resulting models
might be expected to produce more accurate results, with smaller
generalisation errors, and with better out-of-distribution
generalisation.

\begin{figure}
\begin{center}
\includegraphics[width=0.8\textwidth]{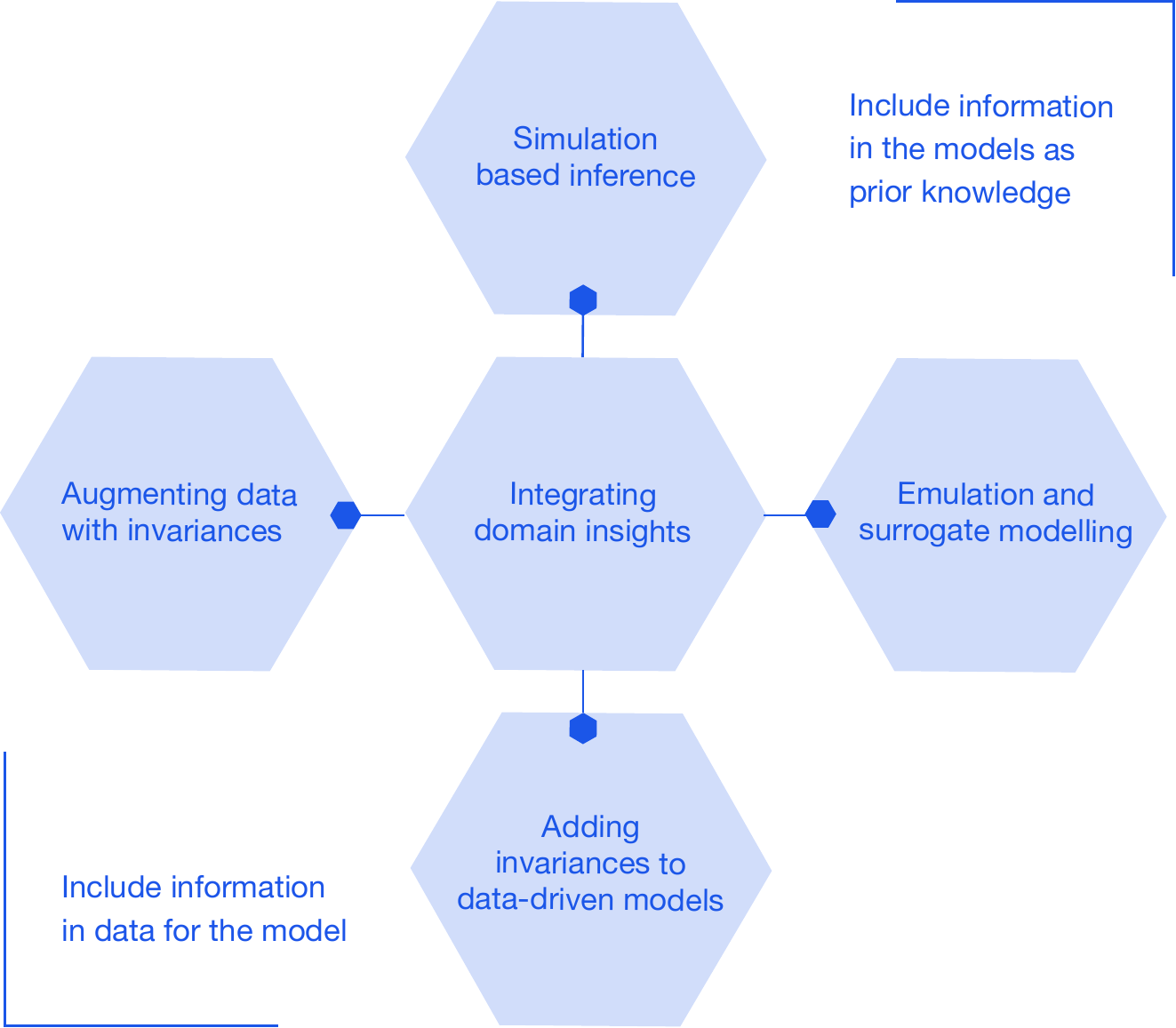}
\end{center}
\caption{Strategies for integrating domain insights: including information in data and including information as prior knowledge.}
\end{figure}

\subsection{Scientific centaurs}\label{scientific-centaurs}

Complementing modelling strategies to encode scientific knowledge are
deployment strategies to use AI in the lab. The lab has long provided a
physical hub for collaboration and knowledge-generation, its function
and form having remained broadly consistent across centuries of
scientific progress. Today, the digitisation of experimental equipment
and laboratory processes offers opportunities to integrate AI in
experimental design and create new virtual labs.

By combining data from measurement devices, simulations of laboratory
processes, and computational models of research or user objectives,
these virtual labs provide a digital sibling of in-person research
activities that can be used to optimise such activities. In drug
discovery, for example, virtual labs could accelerate the testing and
analysis processes that identify candidate drugs from potential drug
targets. Instead of relying on physical testing of such starting
molecules, multiple rounds of virtual testing can rapidly simulate the
processes of drug design, manufacture, testing, and analysis to assess
which starting molecules are more (or less) likely to be viable
candidate drugs \cite{Klami-virtual22}.
As a result, AI can help accelerate the research process.

Advances in machine learning methods to enable effective simulations,
causal modelling, and encoding pre-existing domain insights -- while
packaging such methods into usable toolkits -- are all necessary
foundations for such digital siblings. Moving from virtual laboratory to
`AI assistants' requires further advances in AI system design to create
AI agents that can elicit guidance or input from their domain experts.
Such agents would not only provide useful intuitions for scientific
modelling, but would serve as `scientific sidekicks', actively helping
researchers to drive their research.

This new type of AI assistant would combine the ability to model the
research problem of interest with the ability to model the goals and
preferences of their expert users, even when the user themselves might
not be able to clearly articulate those goals. As a starting point,
these systems would need to support forms of user interaction that can
extract user knowledge, leveraging this to identify appropriate courses
of action. To operate in contexts where user goals might be uncertain
and user behaviour might change in response to the outputs of the AI
system, these AI sidekicks will need insights from cognitive science,
studies of team decision-making, and new learning strategies based on
limited examples. The sophisticated user modelling so-created would
unlock new forms of human-AI collaboration; scientific centaurs that
combine both human and machine intelligence \cite{Celikok-best22}.

\subsection{Enabling communication across
domains}\label{enabling-communication-across-domains}

Underpinning these efforts to integrate pre-existing knowledge into the
design and deployment of AI systems is a feedback loop between domain
and machine learning research, in which each elicits from and feeds into
the other. This loop requires the ability to exchange knowledge and
insights across disciplines through interdisciplinary collaboration and
communication.

Matching model to user need requires shared understandings of the
research question at hand, the constraints -- whether from data,
compute, funding, or time and energy available -- that affect different
collaborators, and the user needs of the domain environment. While AI
researchers might be tempted to develop complex models, showcasing
assorted theoretical and methodological advances in the field, from a
domain perspective, a relatively `simple' model may seem preferable.
Collaborators need to be able to mutually explore what is possible,
while also considering what is useful.

To complete the loop, outputs from machine learning models need to feed
back into the application domain: insights from AI need to be accessible
in ways that allow the transfer of learning from model to user. This
implies some level of explainability. It is not sufficient for an AI
system to produce highly accurate results; those results must also be
interpretable by a domain researcher. As the complexity of AI systems
increases, however, understanding why these systems have produced a
particular result becomes increasingly challenging. While not an issue
for all machine learning methods, this complexity often results in
difficulties explaining the functioning of AI systems.

In response, AI researchers have developed a variety of different
methods to interrogate how AI systems work, or why a particular output
has been produced. Again, to understand which of these methods is
desirable in the context of a scientific application, researchers must
collaborate closely with domain experts. In the context of
pharmaceutical experiments where the aim is to measure how many target
cells are killed off at different dosages of a drug (or drug
combination), for example, researchers might be seeking to `sense-check'
how different drug dosages affect the model, before investigating
specific drugs more rigorously. In astronomical studies, researchers are
often working with high-dimensional datasets with many confounding
correlations. For example, gravitational waves are ripples in space-time
catalysed by the movement of massive bodies in space, such as planets or
stars \cite{NASA-gravitational}.
These invisible phenomena are studied at observatories across the
world,\footnote{See, for example, the LIGO project. Information
  available at:
  \url{https://www.ligo.caltech.edu}}
based on models to describe wave signals and the `noise' generated by
instruments that measure them \cite{Dax-gravitational21}.
Measurements of gravitational waves can be used to infer the properties
of black holes that create them, such as their location, mass, and spin,
using simulation-based inference to characterise the source of a wave,
given the data that detects it. To make such methods more efficient than
existing analytical tools, researchers need to take into account the
structure that sits underneath it: for example, gravitational wave
detectors are located across the globe, and their location affects the
angle at which they detect waves hitting the Earth. This structure can
be exploited through data sampling strategies to help make machine
learning more efficient \cite{Dax-gravitational21}. An alternative, however, is to
use deterministic models that already reflect relevant physical
laws \cite{Bodin-black21}. Across these approaches,
software packages play an important role in enabling communication and
dissemination of methods for wider use.\footnote{See, for example:
  \url{https://lscsoft.docs.ligo.org/bilby/}}

\subsection{Directions}\label{directions-2}

New modelling approaches and mathematical innovations offer exciting
opportunities to integrate domain knowledge, symmetries and invariances
into AI systems \cite{Villar-dimensionless22}.
Integration can be achieved in different ways:

\begin{itemize}
\item
  Data augmentation can help exploit invariances and symmetries,
  resulting in improved model performance, by including in the data
  domain knowledge for a model to ingest.
\item
  Symmetries can be embedded in the design of deep learning systems, for
  example by using the same convolutional filters in different locations
  of an image, CNNs can leverage translation and rotation symmetries.
\item
  Latent force models allow representations of known symmetries
  alongside probabilistic factors, enabling integration of mechanistic
  models with unknown forces \cite{Alvarez-llfm13,Ward-blackbox20}.
\item
  Architectural features can restrict model focus to outputs that
  satisfy symmetries, for example using weight sharing, irreducible
  representations, or invoking symmetries as constraints.\footnote{See,
    for example: \cite{Kondor-generalization18,Maron-invariant18,Dym-universality20}}
\item
  Loss functions can be deployed to penalise predictions that fail to
  satisfy physical constraints or symmetries.
\end{itemize}

In the process, emerging mathematical questions include: how can AI
learn invariances from data? And is it possible to quantify the
performance gain achieved through this?

Research to develop AI assistants in the lab raises interesting
questions about learning strategies and human-machine collaboration.
These AI agents would need to be able to learn how to assist another
agent, in a multi-agent decision-making scenario, where goals might be
unclear, uncertain, or changeable. To tackle this challenge:

\begin{itemize}
\item
  Decision-making with delayed reward or zero-shot learning can help
  agents solve tasks when there is little or nothing known about the
  reward function, and no previous behaviour to learn from.
\item
  Interactive knowledge elicitation \cite{Sundin-improving18},
  combining prior knowledge from cognitive science with learning from
  data \cite{Kangasraasio-parameter19},
  and generative user models \cite{DePeuter-toward21}
  can support more effective interactions between user and machine.
\end{itemize}

Across these areas, care is needed in the design of the points of
interaction between human and AI system. A core question here is: how
can AI researchers extract domain knowledge from relevant experts and
integrate it into a machine learning model? Insights from human-machine
interaction studies and collaborative decision-making systems are
necessary to create effective interfaces between human and machine,
based on factors such as:

\begin{itemize}
\item
  What forms of visualisation are helpful for human users?
\item
  What types of interpretability or explainability are needed for a user
  to achieve their desired interactions?
\item
  What might be the unintended consequences of human-machine
  interaction, such as over-confidence in results or over-reliance on
  the AI system?
\item
  What `theory of mind' is needed to anticipate how human users might be
  likely to respond to an AI system?
\end{itemize}

A challenge in these interactions is that much of the relevant knowledge
held by the domain expert might be qualitative: an intuition of how a
system works, developed over a long period of study, rather than
quantifiable insights.

\subsection{Talks given during this workshop session}

\abstracttitle{Virtual laboratories for science, assisted by collaborative AI}
\abstractauthor[Samuel Kaski]{Samuel Kaski (Aalto University, FI)}
\license

I introduced two ideas: virtual laboratories for science, aiming to introduce an interface between algorithms and domain science that enables AI-driven scale advantages, and AI-based ‘sidekick’ assistants, able to help other agents research their goals, even when they are not able to yet specify the goal explicitly, or it is evolving. Such assistants would ultimately be able to help human domain experts run experiments in the virtual laboratories. I invited researchers to join the virtual laboratory movement, both domain scientists in hosting a virtual laboratory in their field and methods researchers in contributing new methods to virtual laboratories, simply by providing compatible interfaces in their code. For developing the assistants, I introduced the basic problem of agents that are able to help other agents reach their goals, also in zero-short settings, formulated the problem, and introduced solutions in the simplified setting of prior knowledge elicitation, and in AI-assistted decision and design tasks.

\abstracttitle{Making data analysis more like classical physics}
\abstractauthor[David W. Hogg]{David W. Hogg (New York University, US)}
\license

The laws of physics are very structured: They involve coordinate-free forms, they are equivariant to a panoply of group actions, and they can be written entirely in terms of dimensionless, invariant quantities. We find that many existing machine-learning methods can be very straightforwardly modified to obey the rules that physical law must obey; physics structure can be implemented without big engineering efforts. We also find that these modifications often lead to improvements in generalization, including out-of-sample generalization, in natural-science contexts. We have some intuitions about why.

The second example is work by Dan Sheldon on analysis of doppler radar
to extract bird biomass and motion. The radar measures the radial
velocity modulo a constant (i.e., the velocity wraps around to zero).
Previous work had attempted to "unwrap" the data using heuristics. Dan
instead incorporated the modulus operation into the likelihood function
and then developing an algorithm for maximizing this somewhat nasty
likelihood. The result has revolutionized radar analysis and has been
deployed in the BirdCast product from the Cornell Lab of Ornithology.

The third example is the species occupancy model introduced by MacKenzie
et al (2002). When human observers conduct wildlife surveys, they may
fail to detect a species even though the species is present. The
occupancy model combines this detection probability with a habitat
model. However, the expressiveness of the two models (detection and
habitat) must be carefully controlled. Rebecca Hutchinson and I learned
this when we tried to replace the linear logistic regression models with
boosted trees.

In all cases, downstream use of the estimates that come from such data
collection models must be aware of the measurement uncertainties. How
can we correctly quantify those uncertainties and incorporate them in
the downstream analysis? Maybe there are lessons ecologists can learn
from physicists?

\abstracttitle{Latent force models}
\abstractauthor[Mauricio A. \'Alvarez]{Mauricio A. \'Alvarez (University of Manchester, GB)}
\license

A latent force model is a Gaussian process with a covariance function inspired by a differential operator. Such a covariance function is obtained by performing convolution integrals between Green's functions associated with the differential operators, and covariance functions associated with latent functions. Latent force models have been used in several different fields for grey box modelling and Bayesian inversion. In this talk, I will introduce latent force models and several recent works in my group where we have extended this framework to non-linear problems.

\abstracttitle{Translating mechanistic understandings to stochastic models}
\abstractauthor[Carl Henrik Ek]{Carl Henrik Ek (University of Cambridge, GB)}
\license

Statistical learning holds the promise of being the glue that allows us to improve knowledge parametrised explicitly by a mechanistic model with implicit knowledge through empirical evidence. Statistical inference provides a narrative of how to integrate these two sources of information leading to an explanation of the empirical evidence in "light" of the explicit knowledge. While the two sources of knowledge are exchangeable in terms of predictive performance they are not if our focus is that of statistical learning as a tool for science where we want to derive new knowledge.

In this talk we will focus on challenges associated with translating our mechanistic understanding into stochastic models such that they can be integrated with data. In particular, we will focus on the challenges of translating composite knowledge. We will show how these structures and the computational intractabilities they lead to make knowledge discovery challenging. 

The perceived `success' of machine learning comes from application where we have large volumes of data such that only simple and generic models are needed in order to regularise the problem. This means that much of the progress that have been made with predictive models are challenging to translate into useful mechanisms for scientific applications. In this talk we will focus on challenges associated with translating our mechanistic understanding into stochastic models such that they can be integrated with data. In specific we will focus on the challenges of translating composite knowledge. We will show how these structures and the computational intractabilities they lead to makes knowledge discovery challenging. We will discuss properties that we desire from such structures and highlight the large gap that exists with current inference mechanism.

\section{A research agenda in AI for
science}\label{a-research-agenda-in-ai-for-science}

`AI for science' sits at a nexus of disciplines, methods, and
communities. Both AI and `science' (broadly defined) share a core
interest in learning from data. From this interest emerge different
research directions: for AI, questions about the nature of intelligence
and how to understand the learning process in humans and machines; for
science, the outputs of this learning process are the focus, with the
aim of adding new knowledge about natural, physical, and social systems.
A distinctive feature of the emerging `AI for science' agenda is the
ability to move between these worlds, using AI to drive progress in
science and taking inspiration from science to inspire progress in AI.
The result is a continuum of modelling approaches along a spectrum from
strongly mechanistic to statistical models, which allow researchers to
introduce or operate at different levels of abstraction.

The AI for science community therefore combines the ambitions of AI
research with domain-specific goals to advance the frontiers of research
and innovation in their discipline, with an engineering focus on
designing systems that work in deployment, while operating across scales
from the nano- to the interstellar. From these interfaces emerges a
research agenda that --- if successful --- promises to accelerate progress
across disciplines. Inspired by discussions at the Dagstuhl workshop, a
list of research questions arising from this agenda is given in Annex 2.
These span three themes:

\noindent\textbf{Building AI systems for science:} Attempts to deploy AI
in the context of scientific discovery have exposed a collection of gaps
in current machine learning and AI capabilities. Further work is needed
to develop the technical capabilities that will allow AI to be used more
effectively in research and innovation; developing those capabilities
also offers opportunities to contribute to wider attempts to deliver
sophisticated AI systems. Areas for progress include:

\begin{itemize}
\item
  Advancing methods, software and toolkits for high-quality simulation
  and emulation, which integrate effective uncertainty quantification
  and leverage advances in machine learning robustness to ensure they
  operate safely and effectively.
\item
  Detecting scientifically meaningful structure in data, through
  advances in causal machine learning.
\item
  Encoding domain knowledge in AI systems through integration of
  scientific laws, principles, symmetries, or invariances in machine
  learning models, and through virtual, autonomous systems to make
  research more effective.
\end{itemize}

\noindent\textbf{Combining human and machine intelligence:} Effective
deployment of AI in science requires effective interactions between
human, domain and machine intelligence across all stages of the
deployment pathway. AI systems can be made more effective by integrating
pre-existing knowledge about the system of study, but mechanisms are
needed to extract and encode that knowledge. Effective interfaces are
also required in the reverse direction. Translating the outputs of AI
analysis to increased human capability requires an understanding of what
insights are relevant, how they are best communicated, and the cultural
environment that shapes the conduct of science. Areas for progress
include:

\begin{itemize}
\item
  Designing interfaces between humans and machines or AI agents that can
  extract, formalise, and assimilate knowledge that domain researchers
  have acquired, including tacit knowledge, and that communicate new
  knowledge back to the user as actionable insights.
\item
  Building mechanisms for explainability that allow researchers to
  interrogate why and how an AI system delivered a particular result,
  with the explanations provided being tailored to user need.
\item
  Accelerating the pace of knowledge creation and use, through systems
  that mine the existing research knowledge base or that automate
  repetitive or time-consuming elements of the research process.
\end{itemize}

\noindent\textbf{Influencing practice and adoption:} By learning from
recent experiences of deploying AI for science, the field has an
opportunity to promote wider uptake and progress in both scientific
domains and in AI research. This requires capturing both the knowledge
that the community has already generated, about how to design AI
systems, and the know-how about how to overcome practical challenges
that accompanies it, while taking action to grow the community of
researchers excited about the potential of AI in science. Areas for
progress include:

\begin{itemize}
\item
  Supporting new applications, through challenge-led research programmes
  that promote interdisciplinary collaborations and support co-design of
  AI systems to help tackle scientific challenges.
\item
  Developing toolkits and user guides that allow researchers to
  understand which AI tools are suitable for which purposes, and how to
  deploy those tools in practice.
\item
  Sharing skills and know-how, through community outreach that
  disseminates knowledge and know-how in how to use AI.
\end{itemize}

Together, these areas for action highlight the importance of interfaces
-- between researchers and between modelling approaches -- in shaping
the development of AI for science (Figure 3).

\begin{figure}
\begin{center}
\includegraphics[width=\textwidth]{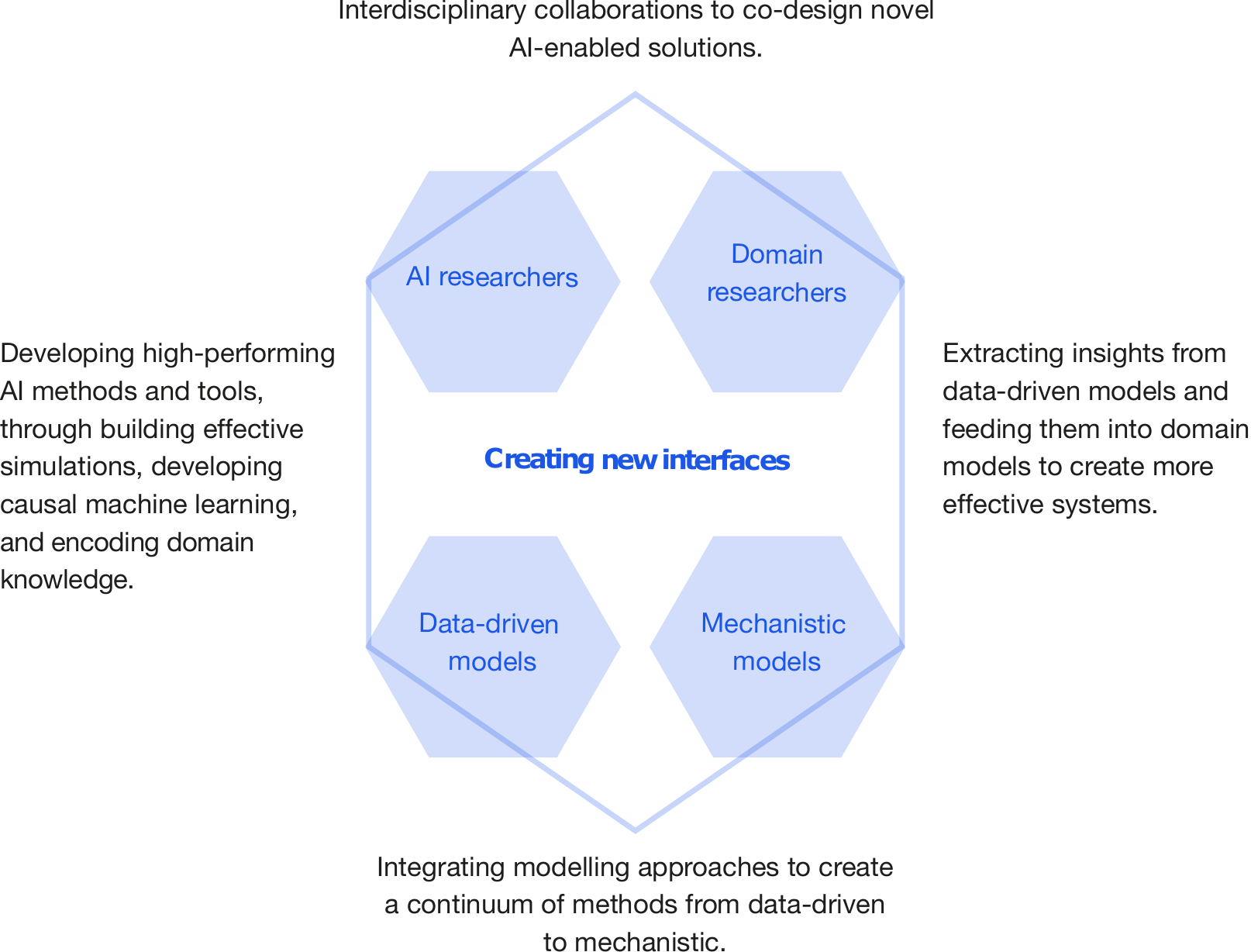}
\caption{Interfaces between machine learning and domain researchers, and between data-driven and mechanistic models.}
\end{center}
\end{figure}

\section{Accelerating progress in AI for science}\label{accelerating-progress-in-ai-for-science}

Building on the impressive advances that machine learning has already
supported in many domains, widespread adoption of AI for research has
the potential to catalyse a new wave of innovations that in turn could
drive greater health, wealth, and wellbeing. The question facing
researchers, funders, and policymakers today is how to harness that
potential. The challenge is to build capability across the research
landscape, connect areas of expertise to areas of need, and to
accelerate the transfer of successful ideas between domains.

The experiences of deploying AI for science described in this document,
and the research agenda that results from these experiences, suggest a
roadmap for action. That roadmap charts a pathway to create an enabling
environment for AI in science, by advancing research that delivers AI
methods to support scientific discovery, building tools and resources to
make AI accessible, championing interdisciplinary research and the
people pursuing it, and nurturing a community at the interface of these
different domains. Progress across these areas can unlock scientific and
methodological advances in AI for science, while also helping answer an
emerging question about whether there exists a core discipline of `AI
for science'. The shared themes and interests that emerge from research
projects at the interface of AI and scientific domains suggest that
there is potential for `AI for science' to surface as a distinct
speciality in computer science. In parallel, domain-specific efforts to
drive the adoption of AI as an enabler of innovation are also needed to
deliver the benefits of AI for scientific discovery.

\subsection{Advance new methods and
applications}\label{advance-new-methods-and-applications}

Efforts to deploy AI in the context of research have highlighted
cross-cutting challenges where further progress in AI methods and theory
is needed to create tools that can be used more reliably and effectively
in the scientific context. Effective simulations are needed to study the
dynamics of complex systems; causal methods to understand why those
dynamics emerge; and integration of domain knowledge to relate those
understandings to the wider world. While elements of these research
challenges are shared with other fields -- topics such as robustness,
explainability, and human-machine interaction also come to the fore in
fields such as AI ethics, for example -- they share an intersection in
the use of AI for science, in the context of efforts to bridge
mechanistic and data-driven modelling.

Alongside these `AI' challenges are a collection of `science'
challenges, where researchers, policymakers and publics have aspirations
for AI to deliver real-world benefits.\footnote{See, for example: the
  EU's Innovation Missions
  \url{https://research-and-innovation.ec.europa.eu/funding/funding-opportunities/funding-programmes-and-open-calls/horizon-europe/eu-missions-horizon-europe_en}
  and UN SDG's
  \url{https://sdgs.un.org/goals}}
Such challenges offer the opportunity to accelerate progress in AI,
while facilitating interdisciplinary exchanges, and opening the field to
input from citizen science or other public engagement initiatives. In
developing these research missions, care is needed to define
cross-cutting questions or challenges that broaden scientific
imaginations, rather than restricting them. The process of converting a
complicated scientific problem into something tractable with AI
necessarily involves some narrowing of focus; to be successful,
mission-led innovation efforts must achieve this focus without losing
meaning, or creating benchmarks that misrepresent the complexity of the
real-world challenge.

Defining shared challenges could help rally the AI for science community
and drive progress in both methods and applications of AI in science.
There already exists examples of how such challenges can build
coalitions of researchers across domains from which the field can draw
inspiration. These include the GREAT08 project, which developed image
analysis techniques to study gravitational lensing \cite{Bridle-great0810};
the Open Problems in Single Cell Biology challenge, which convened the
machine learning community to make progress in Multimodal Single-Cell
Data Integration;\footnote{For further information, see:
  \url{https://openproblems.bio/neurips\_2021/}} and the SENSORIUM challenge,
focused on advancing understandings of how the brain processes visual
inputs.\footnote{For further information, see:
  \url{https://sensorium2022.net/home}} In pursuing this agenda, researchers
can leverage well-established protocols in open-sourcing materials and
sharing documentation to help ensure research advances are rapidly and
effectively disseminated across disciplines. The result should be more
effective methods, and an agile research environment where researchers
can flex methods across disciplines.

\subsection{Invest in tools and
toolkits}\label{invest-in-tools-and-toolkits}

Complementing these efforts to build and share knowledge, well-designed
software tools can help make accessible the craft skills (or know-how)
that make AI for science projects successful. Modelling is a core
component of all AI for science projects. In some aspects, the task for
the field can be thought of as charting a path between the statistician,
whose effectiveness comes from proximity to the domain but whose methods
struggle to scale, and the mathematician, whose tools are adopted across
domains but with some loss of meaning as the distance between
method-generator and adopter increases.

The energy already invested in building effective machine learning
models can be leveraged for wider progress across domains through
investment in toolkits that support the generalisation of effective
approaches. Wide-spectrum modelling tools could offer `off the shelf'
solutions to common AI for science research questions. The challenge for
such toolkits is to create an effective interface between tool and user.
Connecting with the field of human-computer interaction could generate
design insights or protocols to help create more effective human-AI
interfaces.

Best practices in software engineering can help, through documentation
that supports users to successfully deploy modelling tools. User guides
-- or taxonomies of which models are best suited for which purposes and
under what circumstances --- can also help make accessible to non-expect
users the accumulated know-how that machine learning researchers have
gained through years of model development and deployment.

A related engineering challenge is that of data management and
pipeline-building. To interrogate how a model works, why a result was
achieved, or whether an AI system is working effectively, researchers
often benefit from being able to track which data contributed to which
output. The data management best practices that allow such tracking need
to be embedded across AI for science projects. Data management
frameworks -- such as the FAIR data principles --- have already been
developed with the intention of making data more available, and useful,
for research. Further investment is now needed in efforts to implement
those principles in practice.

Investment in these foundational tools and resources can help build
understanding of which AI methods can be used and for what purposes,
lowering the barriers to adopting AI methods across disciplines.

\subsection{Build capability across
disciplines}\label{build-capability-across-disciplines}

Central to progress in both research and toolkit engineering is the
availability of talented researchers with a passion for advancing
science through AI. People matter at all stages of the AI development
and deployment pipeline. Successful projects rely on researchers who are
motivated to work at the interface of different domains; collaborators
who can explain and communicate core concepts in their work across
disciplinary boundaries; engineers who can translate the needs of
different users into AI toolkits; and convenors that can inspire wider
engagement with the AI for science agenda.

Building these capabilities requires multiple points of engagement.
Domain researchers need access to learning and development activities
that allow them to understand and use foundational methods in machine
learning, whether as formal training or through the availability of
tutorials or user guides. AI researchers need access to the scientific
knowledge that should shape the methods they develop, the skills to
translate their advanced knowledge to materials that can be shared for
wider use, and the capacity to dedicate time and resource to learning
about domain needs.\footnote{A comparison here can be drawn with the
  development of statistics as an enabling discipline for many domains:
  statisticians have devoted time to understanding domain practices and
  integrating their work within those practices, often dedicating
  significant resource to understand the nature of the datasets with
  which they are working, before introducing modelling ideas.} Both need
skills in communication, organisation, and convening to operate across
disciplines. Without such capability-building, disciplines risk
remaining siloed; domains developing unrealistic expectations about what
AI can deliver in practice, and AI losing touch with the scientific
questions that are most meaningful to domains.

Institutional incentives shape how individuals engage (or not) with such
interdisciplinary exchanges. Interdisciplinary research often takes
longer and lacks the outlets for recognition available to those working
in single domains, affecting both the motivation of and opportunities
for career progression that are open to those working at the interface
of different disciplines. Much of the engineering work required to make
data and AI accessible beyond a specific project and useful to a wider
community is also traditionally unrecognised by academic incentive
structures. Aligning individual and institutional incentives in support
of interdisciplinarity is a long-standing challenge in research, and one
that becomes more critical to address in the context of developments in
AI. In this context, there may be new opportunities to recognise and
reward successes in AI for science, whether through new fellowships,
prizes, or ways of promoting the work done by those at this interface.

\subsection{Grow communities of research and
practice}\label{grow-communities-of-research-and-practice}

The areas for action described above feed into and from each other.
Progress in research and application can be leveraged to inspire a
generation of researchers to pursue interdisciplinary projects;
effective toolkits can make such progress more likely; skills-building
initiatives can prime researchers to be able to use these toolkits; and
so on, to create an environment where researchers and research advances
transition smoothly across disciplines, leading to a rising AI tide that
lifts all disciplines. Communities of research and practice are the
backdrop for creating such positive feedback loops.

A collection of AI for science initiatives are already building links
across the research landscape. The Machine Learning for Science Cluster
of Excellence at the University of T\"ubingen is leveraging the strength
of its local ecosystem in AI to drive wider progress in research and
innovation;\footnote{Programme website available at:
  \url{https://uni-tuebingen.de/en/research/core-research/cluster-of-excellence-machine-learning/home/}.}
the Accelerate Programme for Scientific Discovery at the University of
Cambridge is building bridges across disciplines, building a community
passionate about opportunities in AI for science;\footnote{Programme
  website available at:
  \url{https://acceleratescience.github.io}.}
the University of Copenhagen's SCIENCE AI Centre provides a focal point
for AI research and education in its Faculty for Science;\footnote{Programme
  website available at:
  \url{https://ai.ku.dk}.} New York
University's Center for Data Science hosts interdisciplinary faculty
pursuing innovative research and education;\footnote{Programme website
  available at: \url{https://cds.nyu.edu}.}
the University of Wisconsin-Madison's American Family Insurance Data
Science Institute is developing strategic partnerships to accelerate the
use of data science in research;\footnote{Programme website available
  at:
  \url{https://datascience.wisc.edu/institute/}.}
new investments by Schmidt Futures across a network of research
institutions are supporting new postdoctoral fellowships at the
interface of AI and sciences \cite{Schmidt-accelerate22}.
Together, these initiatives demonstrate the appetite for progress in AI
for science.

There is an opportunity today to leverage these emerging interests into
a wider movement. Existing initiatives can drive capability-building, by
making training and user guides open, reaching out to engage domain
researchers in skills-building activities, and fostering best practice
in software and data engineering across disciplines. The links they
establish across research domains can form the basis of new
communication channels, whether through discussion forums, research
symposia, or newsletters to share developments at the interface of AI
and science. These communications can be deployed to raise the profile
of people and projects at this interface, celebrating successes, sharing
lessons, and demonstrating the value of interdisciplinary work.
Together, they can help develop an infrastructure for AI in science.

That infrastructure may also benefit from new institutional
interventions to address long-standing challenges in interdisciplinary
AI. New journals could provide an outlet to publish and recognise
high-quality AI for science research, bringing in contributions from
multiple disciplines and helping translate lessons across areas of work.
Membership organisations could help foster a sense of belonging and
community for researchers working at the interface of AI, science, and
engineering, developing career pathways and incentives. Efforts to
convene across disciplines can also catalyse new connections and
collaborations.

Emerging from these efforts is a paradigm shift in how to drive progress
in science. Historically, a small number of foundational texts have been
the catalyst that changed how researchers studied the world; Newton's
Principia; Darwin's Origin of Species; and so on. For much of its modern
history, scientific knowledge has been transmitted through textbooks;
canonical descriptions of the current state of knowledge. Today, the
transformative potential of AI is driven by its pervasiveness; its
impact in science will be achieved through integration across
disciplines. This integration requires widespread mobilisation,
convening machine learning researchers, domain experts, citizen
scientists, and affected communities to shape how AI technologies are
developed and create an amenable environment for their deployment. It
takes a community.

\subsection{AI and science: building the
interface}\label{ai-and-science-building-the-interface}

Advances in AI have disrupted traditional ways of thinking about
modelling in science. Where researchers might previously have
conceptualised models as mechanistic --- reflecting known forces in the
world -- or data-driven, the `AI for science' methods that are emerging
today reject this separation. They are both, combining insights from
mechanistic and data-driven methods, integrating methods to create
something new. What follows from these developments is a spectrum of
modelling approaches, which researchers can deploy flexibly in response
to the research question of interest.

Today, the field of AI for science is characterised by intersections.
Between AI and scientific domains; between science and engineering;
between knowledge and know-how; between human and machine. It operates
across disciplinary boundaries, across scales from the atomic to the
universal, and across both the mission to understand intelligence and
the quest to deploy human intelligence to understand the world. Emerging
from these missions is a continuum of models and methods that allow
researchers to work across domains, extracting the knowledge that humans
have acquired, and levels of inquiry, enhancing that knowledge and
returning it in actionable form.

As both a domain itself and an enabler of other disciplines, the power
of AI in science lies in its ability to convene diverse perspectives in
ways that accelerates progress across research areas. AI for science is
a rendezvous point. Its next wave of development will come from taking
strength from its diversity, and bringing more people into its
community.

\subsection*{Acknowledgments}

The Accelerate Programme for Scientific Discovery would like to thank
Schmidt Futures for its continuing support, and the donation that
enables its work.

\bibliography{ai-for-science}

\appendix
\section{Participants at Machine Learning for Science: Bridging
Data-driven and Mechanistic Modelling (Dagstuhl Seminar 22382, 18-23
September
2022)}\label{annex-1-participants-at-machine-learning-for-science-bridging-data-driven-and-mechanistic-modelling-dagstuhl-seminar-22382-18-23-september-2022}

\emph{Organisers: Philipp Berens, Kyle Cranmer, Neil D. Lawrence, Jessica
Montgomery, Ulrike von Luxburg.}

Thank you to all those who contributed to Dagstuhl Seminar 22382,
discussions at which are the foundation for this paper:

\begin{itemize}
    \item 
    Mauricio A Álvarez (University of Manchester, GB)
    \item
    Bubacarr Bah (AIMS South Africa -- Cape Town, ZA)
    \item
    Jessica Beasley (Collective Next -- Boston, US)
    \item
    Philipp Berens (Universität Tübingen, DE)
    \item
    Maren Büttner (Helmholtz Zentrum München \& Universität Bonn)
    \item
    Kyle Cranmer (University of Wisconsin -- Madison, US)
    \item
    Thomas G. Dietterich (Oregon State University -- Corvallis, US)
    \item
    Carl Henrik Ek (University of Cambridge, GB)
    \item
    Stuart Feldman (Schmidt Futures -- New York, US)
    \item
    Asja Fischer (Ruhr-Universität Bochum, DE)
    \item
    Philipp Hennig (Universität Tübingen, DE)
    \item
    David W. Hogg (New York University, US)
    \item
    Christian Igel (University of Copenhagen, DK)
    \item
    Samuel Kaski (Aalto University, FI)
    \item
    Ieva Kazlauskaite (University of Cambridge, GB)
    \item
    Hans Kersting (INRIA -- Paris, FR)
    \item
    Niki Kilbertus (TU München, DE \& Helmholtz AI München, DE)
    \item
    Neil D. Lawrence (University of Cambridge, GB)
    \item
    Gilles Louppe (University of Liège, BE)
    \item
    Jakob Macke (Universität Tübingen, DE)
    \item
    Dina Machuve (DevData Analytics -- A, TZ)
    \item
    Eric Meissner (University of Cambridge, GB)
    \item
    Siddharth Mishra-Sharma (MIT -- Cambridge, US)
    \item
    Jessica Montgomery (University of Cambridge, GB)
    \item
    Jonas Peters (University of Copenhagen, DK)
    \item
    Aditya Ravuri (University of Cambridge, GB)
    \item
    Markus Reichstein (MPI für Biogeochemistry -- Jena, DE)
    \item
    Bernhard Schölkopf (MPI für Intelligente Systeme -- Tübingen, DE)
    \item
    Francisco Vargas (University of Cambridge, GB)
    \item
    Soledad Villar (Johns Hopkins University -- Baltimore, US)
    \item
    Ulrike von Luxburg (Universität Tübingen, DE)
    \item
    Verena Wolf (Universität des Saarlandes -- Saarbrücken, DE)

\end{itemize}

\section{Research
questions arising from the `AI for science research agenda' discussion
during the Dagstuhl workshop}

\subsection*{Building AI systems for science}

\begin{itemize}
\item
  How can AI systems accurately generalise from finite observations? How
  can they detect causality or structure from finite observations?
\item
  What is the computational cost of complexity, and what methods can
  help manage this?
\item
  What forms of system calibration and uncertainty quantification are
  useful in the context of scientific discovery? Are theoretical
  guarantees necessary?
\item
  What new forms of explainability or interpretability could facilitate
  the deployment of AI in science?
\item
  How could AI support generalisation from a small number of
  observations? What methods could enable few- or one-shot learning?
\item
  How can AI researchers build meaningful models from data to accurately
  represent causal mechanisms in the system of study? How can
  researchers identify the most effective model for their system of
  study?
\item
  What does it mean to understand a model? How can researchers combine
  explainability with complexity?
\item
  How can AI methods be made robust and easy to use in deployment by
  domain scientists?
\item
  How can advances in simulation methods be applied in domains where the
  system at hand is less easily described by equations?
\item
  What advances are needed to expand the use of simulations in science?
  How can AI help simulate laboratory experiments or environments,
  helping make more efficient different elements of the scientific
  process? How might this be expanding in the long-term, for example to
  planning experimental design or helping identify where data is
  missing?
\item
  How can `digital siblings' be used to explore the impact of different
  interventions on complex systems?
\end{itemize}

\subsection*{Combining human and machine intelligence}

\begin{itemize}
\item
  How can AI researchers best extract, formalise and assimilate the
  knowledge that domain researchers have acquired? What forms of
  knowledge representation can formalise scientific understandings of
  the world, translating these to objective functions for AI systems?
  What forms of human-AI engagement can make use of the `qualitative'
  knowledge -- or intuitions about a system -- that domain researchers
  have accumulated?
\item
  How can AI capture the qualitative understanding that researchers have
  of their domain to more accurately or effectively characterise a
  system?
\item
  How can AI be effectively deployed to mine the existing research
  knowledge base -- for example, papers, databases, and so on -- to
  extract new insights?
\item
  Where can automation support research progress? Which elements of the
  scientific process could be automated, and where is human input vital?
\item
  What forms of collaboration are needed to effectively specify helpful
  outputs from an AI system?
\item
  How can insights from AI analysis be returned to researchers in an
  actionable way? What mix of AI design, engineering, social
  interaction, and education can make effective interfaces between
  domain researchers and AI systems?
\item
  How can the outputs of AI systems be made interpretable for scientific
  users?
\item
  How can AI researchers better understand and design for the forms of
  interpretability that resonate with domain researchers?
\item
  What processes of collaboration or co-design can help describe what
  scientists `need to know' from an AI system?
\item
  What best practices or methods can be deployed to effectively
  communicate uncertainty from AI systems to human users?
\end{itemize}

\subsection{Influencing practice and adoption}

\begin{itemize}
\item
  What are the craft skills in AI for science? What `know-how' is
  necessary to make AI work effectively in practice?
\item
  What skills-building or forms of outreach can help take AI tools out
  of the AI community and into `the lab'?
\item
  How has machine learning been used most effectively for research and
  innovation? What best practices, or lessons, do existing efforts in AI
  for science offer?
\item
  Which AI tools are suitable for which purposes, disciplines, or
  experimental designs? Is it possible to create a taxonomy for science?
\item
  Are there generalisable methods or conclusions that can be taken from
  domain-specific efforts to deploy AI for science?
\end{itemize}

\end{document}